\documentclass[11pt,a4paper]{article}
\usepackage[hyperref]{emnlp2020}
\usepackage{times}
\usepackage{latexsym}
\usepackage{amssymb,amsmath,graphicx}
\usepackage{soul}
\usepackage{tikz}
\usepackage{color}
\usepackage{mathrsfs}
\usepackage{algorithm}
\usepackage{algpseudocode}
\usepackage{multirow}
\usepackage{subfigure}
\usepackage{graphicx}
\usepackage{textcomp}
\usepackage{amssymb}
\usepackage{microtype}

\usepackage{times}
\usepackage{latexsym}
\usepackage{amsmath}
\usepackage{pifont}

\usepackage{graphicx}

\usepackage{booktabs}
\usepackage{multirow}
\usepackage{tikz}
\usepackage{amssymb}
\usepackage{textcomp}
\usepackage{colortbl}

\newcommand{\cmark}{\ding{51}}
\newcommand{\xmark}{\ding{55}}

\colorlet{soulred}{red!30}
\DeclareRobustCommand{\hlred}[1]{{\sethlcolor{soulred}\hl{#1}}}

\colorlet{soulbleu}{cyan!20}
\DeclareRobustCommand{\hlblue}[1]{{\sethlcolor{soulbleu}\hl{#1}}}

\colorlet{soulgreen}{green!20}
\DeclareRobustCommand{\hlgreen}[1]{{\sethlcolor{soulgreen}\hl{#1}}}

\colorlet{soulyellow}{yellow!40}
\DeclareRobustCommand{\hlyellow}[1]{{\sethlcolor{soulyellow}\hl{#1}}}

\colorlet{soulorange}{orange!30}
\DeclareRobustCommand{\hlorange}[1]{{\sethlcolor{soulorange}\hl{#1}}}

\colorlet{soulpurple}{blue!30}
\DeclareRobustCommand{\hlpurple}[1]{{\sethlcolor{soulpurple}\hl{#1}}}

\usepackage{microtype}

\aclfinalcopy 

\title{Learning Knowledge Bases with Parameters \\ for Task-Oriented Dialogue Systems}

\author{Andrea Madotto, Samuel Cahyawijaya, Genta Indra Winata, Yan Xu, \\\textbf{Zihan Liu}, \textbf{Zhaojiang Lin}, \textbf{Pascale Fung}\\
Center for Artificial Intelligence Research (CAiRE)\\
  Department of Electronic and Computer Engineering\\
  The Hong Kong University of Science and Technology, Clear Water Bay, Hong Kong\\
  \small\texttt{\{amadotto, scahyawijaya, giwinata, yxucb, zliucr, zlinao\}@connect.ust.hk},\\ \small\texttt{pascale@ece.ust.hk}}

\date{}

\begin{document}
\maketitle
\begin{abstract}
Task-oriented dialogue systems are either modularized with separate dialogue state tracking (DST) and management steps or end-to-end trainable. In either case, the knowledge base (KB) plays an essential role in fulfilling user requests. Modularized systems rely on DST to interact with the KB, which is expensive in terms of annotation and inference time. End-to-end systems use the KB directly as input, but they cannot scale when the KB is larger than a few hundred entries. In this paper, we propose a method to embed the KB, of any size, directly into the model parameters. The resulting model does not require any DST or template responses, nor the KB as input, and it can dynamically update its KB via fine-tuning. We evaluate our solution in five task-oriented dialogue datasets with small, medium, and large KB size. Our experiments show that end-to-end models can effectively embed knowledge bases in their parameters and achieve competitive performance in all evaluated datasets\footnote{Code available in \url{https://github.com/HLTCHKUST/ke-dialogue}}.
%


\end{abstract}

\section{Introduction}
Task-oriented dialogue systems are designed to help users achieve predefined goals, such as booking restaurants or movie recommendations via natural language interactions. These systems are deeply connected with external Knowledge Bases (KBs) since the system responses are guided by the output from the KB and the dialogue history.

\begin{figure}[t]
    \centering
    \includegraphics[width=0.98\linewidth]{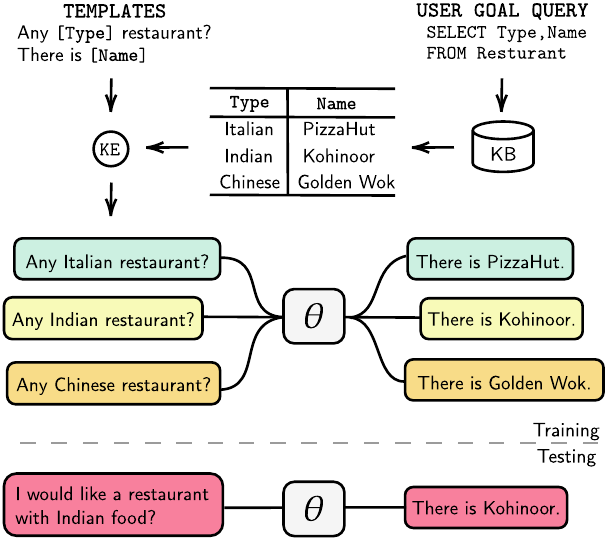}
    \caption{During training, the KE dialogues are generated by fulfilling the \texttt{TEMPLATE} with the \textit{user goal query} results, and they are used to embed the KB into the model parameter $\theta$. At testing time, the model does not use any external knowledge to generate the correct responses.}
    \label{fig:overview}
\end{figure}
The current state-of-the-arts~\cite{lei2018sequicity,zhang2019task,mehri2019structured,chen2019semantically,peng2020soloist,hosseini2020simple} are end-to-end \textit{pipelined} systems that rely on Dialogue State Tracking (DST) and Speech Act (S-ACT) annotations. Aside from the annotation cost, which is knowingly high~\cite{budzianowski2018multiwoz}, these pipelined systems must predict a valid DST for querying the KB, execute the query, generate a response template, and finally fulfill it with the retrieved information. The resulting systems are usually overly complicated, and they require multiple steps, including a direct interaction with the KB. 

On the other end of the spectrum, there are \textit{end-to-end} trainable models that use both the KB and the dialogue history as input, and they directly generate system responses. Most of the implementations use either the \textit{Gold KB} as input~\cite{ericKVR2017,madotto2018mem2seq,qin2019entity,qin2020dynamic,banerjee2019graph,neelakantan2019neural} or an intermediate API call to retrieve part of the KB (\textit{API+KB})~\cite{bordes2016learning,eric-manning:2017:EACLshort,madotto2018mem2seq,reddy2019multi,wu2019global}. These systems require at least the DST annotation for generating the API calls or to select the gold KB. Moreover, even with the most advanced transformer architecture~\cite{kitaev2020reformer,lample2019large,child2019generating}, end-to-end models struggle when the input becomes too large~\cite{neelakantan2019neural}. For example, in MWOZ~\cite{budzianowski2018multiwoz}, there are 22K entities just for one of the domains. Interested readers can refer to Appendix C for an overview of different task-oriented methodologies.

On the other hand, \citet{petroni2019language} discovered a simple yet effective way to query factual knowledge from BERT~\cite{devlin2019bert}. Later on, \citet{roberts2020much} fine-tuned a pre-trained language model, T5~\cite{raffel2019exploring}, on just question-answers pairs, without letting the model access any external context or knowledge. These results suggest that the actual knowledge is stored in the model parameters. However, in task-oriented dialogue systems, KB entities do not appear in news articles or Wikipedia, e.g., hotel addresses or postcodes, and thus the aforementioned methods cannot be straightforwardly applied, especially when the KB dynamically changes (e.g., weather information).

In this paper, we propose a method to store the KB directly into the model parameters using a novel Knowledge Embedded (KE) approach. The resulting model does not use any DST or template responses, nor a KB as input at the inference time, and it can be used in dynamically changing KBs via fine-tuning. The KE approach consists of a newly defined \textit{user goal query} that generates equivalents KE dialogues from the KB (i.e., table or graph) using minimal annotation effort. Figure~\ref{fig:overview} shows a high level overview of our approach. To verify the effectiveness of our proposed methodology, we extensively experiment, using both automatic and human metrics, in five task-oriented datasets with small, medium, and large KBs. Our experiments show that end-to-end models can effectively embed knowledge bases in their parameters and achieve competitive performance in all five datasets. 

\section{Methodology}
\label{sec:methodology}
In this section, we formalize the Knowledge Embedded (KE) strategy and the learning algorithm. In Section~\ref{subsection:preliminari}, we provide several preliminary definitions used thought out the paper. In Section~\ref{subsection:goal}, we extend the user goal definition from \citet{schatzmann2007agenda} to cover a broad concept that we define as \textit{user goal query}. Then, in Section~\ref{subsection:KE}, we describe two functions, \texttt{KE-DELEX} and \texttt{KE-RELEX}, used for generating \texttt{TEMPLATE}s and KE dialogues, respectively. Finally, in Section~\ref{sec:causal_language_model}, we describe the Causal Language Model Transformer~\cite{vaswani2017attention} used for modeling the dialogue responses. 

\subsection{Preliminary Definition}
\label{subsection:preliminari}
We define a dataset as a set of dialogues $\mathscr{D} = \{\mathcal{D}^1, \mathcal{D}^2, \dots,\mathcal{D}^n\}$. A dialogue $\mathcal{D}$ is a collection of one or more alternating turns between two speakers, such as $\mathcal{D}=\{U_1, S_1, \dots,U_{t},S_{t}\}$, where each $U$ and $S$ are sequences of words. Then, we define a table-formatted KB as a set of tuples $\mathcal{K}=\{ (v_1^{a_1}, \dots v_1^{a_k}), \dots, (v_p^{a_1}, \dots v_p^{a_k}) \}$, where $a_1, \dots a_k \in A$ are the column names of the table, $v_i^{a_j} \in V_{a_j}$ is the value of tuple $i$ for the column name $a_j$, and $V_{a_j}$ is a set of possible values for the column name $a_j$ available in the ontology. 

Following the notation in~\citet{moon2019opendialkg}, we define a graph-formatted KB as $\mathcal{G} = \mathbf{N}_{KG}\times \mathbf{R}_{KG}$, where $\mathbf{N}_{KG}$ and $\mathbf{R}_{KG}$ are the nodes and the relation set, respectively. Then, we define $\mathbf{N}_r(n)$ as a set of directly connected neighbours of $n\in \mathbf{N}_{KG}$ by a relation $r\in \mathbf{R}_{KG}$. Similarly, we define $\mathbf{N}_{Rh}(n)$ to be a set of nodes connected to $n$ via $h$-hops with a set of relations $R$. 

\begin{table*}[!t]
\centering
\resizebox{0.99\textwidth}{!}{
\begin{tabular}{l|c|c|c|c|llllll}
\hline
\multicolumn{6}{c}{\textbf{\textbf{User Goal Query}}} & \multicolumn{5}{c}{\textbf{\texttt{TEMPLATE}}} \\ \hline
\multicolumn{6}{l}{\texttt{SELECT} \hlblue{type}, \hlgreen{poi}, \hlorange{distance}, \hlred{address}} & \multicolumn{5}{l}{\textit{U:} Where is the closest \hlblue{
[type]}?} \\
\multicolumn{6}{l}{\texttt{FROM} navigation} & \multicolumn{5}{l}{\textit{S:} \hlgreen{[poi]} is \hlorange{[distance]} away} \\
\multicolumn{6}{l}{\texttt{GROUP BY} \hlblue{type}} & \multicolumn{5}{l}{\textit{U:} What is the address?} \\
\multicolumn{6}{l}{\texttt{HAVING} \hlorange{distance} = \texttt{MIN}(\hlorange{distance})} & \multicolumn{5}{l}{\textit{S:} \hlgreen{[poi]} is located at \hlred{[address]}.} \\ \hline \hline
\multicolumn{6}{c}{\textbf{\textbf{Query Results}}} & \multicolumn{5}{c}{\textbf{\textbf{KE Dialogue}}} \\ \hline
 & \cellcolor[rgb]{0.776,0.937,0.9982}\textbf{type} & \cellcolor[rgb]{0.8,1,0.8}\textbf{poi} & \cellcolor[rgb]{1,0.8509,0.698}\textbf{distance} & \cellcolor[rgb]{1,0.698,0.698}\textbf{address} & & \multicolumn{5}{l}{\textit{U:} Where is the closest \hlblue{gas station?}} \\ \cline{2-5}
 & gas station & Valero & 5 miles &  91 el camino real & &
\multicolumn{5}{l}{\textit{S:} \hlgreen{Valero} is \hlorange{3 miles} away} \\
 & grocery store & safeway & 4 miles & 452 arcadia pl & & \multicolumn{5}{l}{\textit{U:} What is the address?} \\
 & restaurant & pizzahut & 3 miles &  915 arbol dr & & \multicolumn{5}{l}{\textit{S:} \hlgreen{Valero} is located at \hlred{200 Alester Avenue}.} \\
 \hline
\end{tabular}
}
\caption{A sample of the generated Knowledge Embedded (KE) dialogues. The KE Dialogue are generated by fulfilling the \texttt{TEMPLATE}s with the user goal query results.}
\label{tab:delex}
\end{table*}
\subsection{User Goal Query}
\label{subsection:goal}
In task-oriented dialogue systems, the user goal~\cite{schatzmann2007agenda} for a given dialogue $\mathcal{D}$ is defined as $G=(C,R)$, where $C$ is a set of constraints that specify the required information, and $R$ denotes the actual pieces of information of the user desire, (e.g., the name, address, phone number, etc.). The constraint $C$ is usually expressed by specific values for the attribute, e.g., \texttt{\{loc=center,price=cheap\}}, since there is a one-to-one connection between the user goal and the dialogue. In this paper, we hypothesize that by changing the values of the attributes in $C$ (e.g., \texttt{loc=north}) we can generate an equivalent dialogue covering different knowledge.

We leverage the expressive power of \textit{query languages} to describe all the equivalent values that match a particular dialogue, and we name this \textit{User Goal Query}. We use the \texttt{SQL} syntax~\cite{chamberlin1974sequel} for the table-formatted KB and \texttt{CYPHER} syntax~\cite{webber2012programmatic} for the graph-formatted KB.
Following \cite{schatzmann2007agenda}, we define a set of constraints $C$, and requirements $R$ for dialogues with a table-formatted KB, as follows: 
\begin{align}
    C &= \{ \texttt{OP}(a,v) | a\in A, v\in V_a  \}, \\
    R &= \{ a | a\in A\}  \cup \{a | a \in C\},
\end{align}
where \texttt{OP} is the database operation expressable in an \texttt{SQL} query (e.g., ==, MIN, MAX, SUM, AVG, etc.). The user goal query is then written directly as $\texttt{SELECT} \ R \ \texttt{FROM} \ \mathcal{K} \ \texttt{WHERE}\ C.$\footnote{Notice that we include the attribute specified in $C$ into $R$ by overloading the definition of $\in$} 

Similarly, we extend the user goal query definition for datasets with graph-KBs (e.g., OpenDialKG~\cite{moon2019opendialkg}). Let us define the $C$ and $R$ for dialogues with a graph-formatted KB as:
\begin{align}
    C &= \{ r | r\in \mathbf{R}_{KG}\}, \\
    R &= \{ n | \exists \hat n\in \mathbf{N}_{KG}, \hat n\in \mathbf{N}_{rh}(n),r \in C \}, \label{KB_PATH}
\end{align}
where $h$ is the number of hops. The corresponding user goal query is written directly using \texttt{CYPHER} as $\texttt{MATCH} \ C \ \texttt{RETURN}\ R$, where the node in $R$ and $C$ are specified with placeholders (Table A3 in Appendix A). Indeed, a \texttt{CYPHER} query is specified by a graph pattern made of relations in $\mathbf{R}_{KG}$. The query results are nodes connected by the specified pattern. In Appendix A.1, we briefly explain the \texttt{CYPHER} query syntax in more details. 

\begin{table*}[!t]
\centering
\resizebox{0.99\textwidth}{!}{
\begin{tabular}{r|cc|ccc|cc}
\hline
\multicolumn{1}{c|}{\textbf{}} & \multicolumn{2}{c|}{\textbf{Statistics}} & \multicolumn{3}{c|}{\textbf{Seq. Length}} & \multicolumn{2}{c}{\textbf{KE Statistics}}  \\ \hline
\textbf{Name} & \textit{\#\textbf{Dial.}} & \textit{\#\textbf{Utt.}} & \textit{\textbf{Dial.}} & \textit{\textbf{+GoldKB}} & \textit{\textbf{+FullKB}} & \textit{\#\textbf{Temp.}} & \textit{\#\textbf{KE-Dial.}} \\ \hline
\textit{bAbI-5}~\cite{bordes2016learning} & 3,000 & 26,326 & 236 & 347 & 10,236 & 100 & 55,800 \\ \hline
\textit{CamRest}~\cite{wen2016network} & 676 & 2,744 & 156 & 393 & 1,356 & 161 & 32,361 \\ \hline
\textit{SMD}~\cite{ericKVR2017} & 3,031 & 15,928 & 109 & 435 & - & 300 & 2,420 \\ \hline
\textit{MWOZ}$^{\dagger}$~\cite{budzianowski2018multiwoz} & 2,877 & 19,870 & 730 & 996$^{\ddagger}$ & 23,730 & 527 & 58,440 \\ \hline 
\textit{OpenDIALKG}~\cite{moon2019opendialkg} & 15,673 & 91,209 & 225 & 292 & 590,225 & 11,041 & 12,593 \\ \hline
\end{tabular}
}
\caption{Datasets statistics. \textit{\#\textbf{\textit{Temp.}}} indicates the number of the extracted valid \texttt{TEMPLATEs}, \textbf{\#\textit{KE-Dial.}} indicates the number of generated knowledge-embedded dialogues. We count the maximum input lengths for: dialogue-only (\textbf{\textit{Dial.}}), dialogue with golden KB (\textbf{\textit{Dial.+GoldKB}}), and dialogue with full KB (\textbf{\textit{Dial.+FullKB}}). $^{\ddagger}$ as provided by \citet{qin2020dynamic}. $^{\dagger}$ We consider only single domain dialogues. 
}
\label{tab:stat}
\end{table*}
\subsection{Knowledge Embedded (KE)}
\label{subsection:KE}
Given a dialogue $\mathcal{D}$ and the user goal query, we define two functions: \texttt{KE-DELEX} and \texttt{KE-RELEX}. The \texttt{KE-DELEX} is used to generate the dialogue \texttt{TEMPLATE}s, which is a version of $\mathcal{D}$ where the set of entities related to the user goal query is replaced by their corresponding attribute placeholder. We denote with $B$ the dictionary that contains the bidirectional mapping between the entities and the corresponding attribute placeholder. Then, the \texttt{KE-RELEX} uses the results from the user goal query to assign new equivalent values to the placeholder in $B$. Practically, every \texttt{TEMPLATE} generates as many dialogues as the cardinality of the tuples, or the paths, returned by the user goal query. We denote with $\mathscr{D}^{N}$ the newly generated dialogues and we refer to it as KE dialogues. 

For example in Table~\ref{tab:delex}, we show a \texttt{TEMPLATE} and user goal query in the \texttt{SQL} syntax, with its resulting output tuples. The dialogue in the example is generated by \texttt{KE-RELEX} using the first tuple, e.g., \texttt{[Type]} is converted into ``gas station'', \texttt{[poi]} into ``Valero'', and so on.

In the current version of the algorithms, the functions \texttt{KE-DELEX} and \texttt{KE-RELEX} are implemented using string matching. However, they can be implemented using statistical methods; for example, \citet{moon2019opendialkg} proposed a model to generate the graph path given a dialogue. 

\subsection{Causal Language Modeling}
\label{sec:causal_language_model}
In this paper, we model the dialogue responses using a Transformer~\cite{vaswani2017attention}-based Language Model (LM)~\cite{radford2019language} by using the dialogue history as the prefix in $\mathcal{D}$ and by auto-regressively generating the responses word-by-word $S_t$~\cite{DBLP:journals/corr/abs-1901-08149,zhang2019dialogpt}. Let us define the words in $S_t$ as a set \{$s_{1},\dots,s_{n}$\}, then we factorize the language model distribution using the chain rule of probability~\cite{bengio2003neural} as:
\begin{equation}
    p_{\theta}(S_t|\mathcal{D}_t) =  \prod_{i}^{n}p_{\theta}(s_i|s_{<i},\mathcal{D}_t),
\end{equation}
where $\theta$ are the model parameters and $\mathcal{D}_t=\{U_1, S_1, \dots,U_{t}\}$ is the dialogue history. The parameters in $\theta$ are trained to minimize the negative
log-likelihood over a dataset of dialogues $\mathscr{D}$. Formally, we define the $\mathcal{L}$ as following:
\begin{equation}
  \mathcal{L}(\mathscr{D}) =  - \sum_k^{|\mathscr{D}|} \sum_{i}^{n} \log p(s^k_i|s^k_{<i},\mathcal{D}^k_t), 
  \label{eq:loss}
\end{equation}
where $n$ is a maximum response length. Hence, to embed the KB into $\theta$, we include the KE dialogues $\mathscr{D}^{N}$ in the training set, and we train a Transformer-based Language Model with Equation~\ref{eq:loss}.

\section{Experiments}
In all experiments, if not specifically mentioned, we use the pre-trained GPT2 (small)~\cite{radford2019language} as Causal Language Model~\cite{wolf2019transfertransfo}. When the dataset has a sufficiently small KB (i.e., less than 1024 tokens), we also fine-tune GPT2 using the KB as input. In Appendix D, we report details about hyperparameters and the implementation details.  In Appendix E, we report the data splitting for each dataset.

\subsection{Datasets} 
\label{sec:dataset}
We use five publicly available multi-turn task-oriented dialog datasets to evaluate our methodology: bAbI-dialogue (bAbI-5) \cite{bordes2016learning}, Cambridge Restaurant 626 (CamRest)~\cite{wen2016network}, In-Car Assistant (SMD)~\cite{ericKVR2017}, MultiWoZ single (MWOZ)~\cite{budzianowski2018multiwoz}, and OpenDialKG~\cite{moon2019opendialkg}. In all datasets, we use the provided split for train/valid/test, except for OpenDialKG where the split was not provided. Dataset statistics are reported in Table~\ref{tab:stat}, including the sequence length of different settings and the number of \texttt{TEMPLATEs} used for the KE-dialogues.

In all datasets, we use plain text as the input/output sequences instead of their delexicalized version. This makes the task more challenging, but at the same time more practical because the model produces real entities rather than predefined placeholders, and we do not require additional relexicalization step at the inference time. 



\subsection{Evaluation Metrics}
In bAbI, since it is a synthetic dataset, we use the response and dialogue accuracy~\cite{bordes2016learning}. In CamRest, SMD, MWoZ, and OpenDialKG, we use both the BLEU score~\cite{papineni-etal-2002-bleu} and entity F1-score~\cite{ericKVR2017}. In both CamRest and MWOZ, the existing scorer for the Inform and Success rate~\cite{budzianowski2018multiwoz} requires template responses and the predicted DST. Since neither of the two is available for end-to-end models, we implement a plain text scorer for the Inform and Success rate, and we release it, together with our code, for future research. Finally, in OpenDialKG we use the 2-hop neighbors of the entity appearing in the user turn as the gold-reference for the F1-score, which are defined as $\mathbf{N}_{r2}(n) \ \forall{n}\in E(U_t), \exists r \in \mathbf{R} $, where $E(U_t)$ are the list of entity nodes appearing in $U_t$.  

Additionally, we conduct a human evaluation to measure the \textit{Humanness} and \textit{Correctness} of the generated responses. The correctness is computed by counting the ratio of correct entities provided in the generated responses. For the humanness, we use a 4-point Likert Scale, where 1 indicates a non-human-like response, and 4 indicates a very human-like response. All the reported human evaluation results are statistically significant with a p-value$<0.05$. Appendix B provides more details of the human evaluation.

\subsection{Results}
\label{sec:result} 
In this section, we describe baselines, training settings, and \texttt{KE-DELEX} function in each dataset. Table~\ref{tab:stat} summarizes the number of \texttt{TEMPLATEs} and KE dialogues generated in each dataset. All generated \texttt{TEMPLATEs} are extracted from the training dialogues provided in each dataset. More detailed results for all datasets can be found in Appendix F. 

\paragraph{bAbI-dialog} is a synthetic dataset with five sub-tasks for end-to-end task-oriented models~\cite{bordes2016learning}. Task 1 to 4 is about
API calls, refining API calls, recommending options, and providing additional information, respectively. Task 5 is the union of tasks 1-4. Two test-set are provided, one with API combinations appearing in the training set and one with Out-of-Vocabulary APIs. In this paper, we evaluate using task 5 only, in both test sets, by removing all API calls and KB information from the dialogues. 

This dataset provides the user goal query directly, and since it is synthetic, the \texttt{KE-DELEX} function is implemented using a string matching. Moreover, we train a GPT2 from scratch using a word-level tokenizer with the bAbI vocabulary. Table~\ref{tab:babi-results} compares the performance of GPT2, with and without KE, to existing models that use both API and KB as input. As expected, training GPT2 just on the training dialogues,  which covers only 50\% of the KB, does not perform well. Instead, by using the KE dialogues in training, GPT2 consistently generates the correct response in both test sets.

\begin{table}[t]
\centering
    \resizebox{0.45\textwidth}{!}{
        \begin{tabular}{r|c|c}
        \hline
        \multirow{1}{*}{\textbf{Model}} & \multicolumn{1}{c|}{\textbf{Test}} & \multicolumn{1}{c}{\textbf{Test OOV}} \\ \hline
        QRN$^1$ & 99.60 (-) & 67.80 (-) \\
        Mem2Seq$^2$ & 97.90 (69.60) & 84.50 (2.30)  \\
        BoSsNet$^3$ & 97.30 (65.60) & 91.70 (18.50)\\
        GLMP$^4$ & 99.20 (88.50) & 92.00 (21.70)  \\ \hline
        GPT2 & 90.74 (31.00) & 70.14 (0.00) \\ 
        GPT2+KE & \textbf{99.99} (\textbf{99.90}) & \textbf{99.01} (\textbf{94.90}) \\ \hline
        \end{tabular}
    }
    \caption{Results on the bAbI dataset.$^1$~\cite{seo2017query}, $^2$~\cite{madotto2018mem2seq}, $^3$~\cite{raghu2019disentangling}, $^4$~\cite{wu2019global}.}
    \label{tab:babi-results}
\end{table}


\paragraph{CAMREST} is a human-to-human collected dataset for restaurant booking~\cite{wen2016network}. This dataset provides the user goal query, and the \texttt{KE-DELEX} function is implemented using a string matching. We extracted 161 valid \texttt{TEMPLATEs} for a total number of 32,361 KE dialogues. Table~\ref{tab:camrest-results} compares the performance of GPT2, with and without KE, and other models on both automatic and human evaluation. MLMN~\cite{reddy2019multi} and BoSsNet~\cite{raghu2019disentangling} use intermediate APIs to select a subset of the KB, where instead KBRet~\cite{qin2019entity} uses directly the gold KB. To the best of our knowledge, no models used the entire KB as input, thus we train GPT2 using intermediate API and KB. In general, this setting (GPT2+KB) does not perform as well as similar baselines. This because the KB format is very different from the plain text used for the pre-training. Instead, GPT2+KE is able to achieve better performance than the current state-of-the-art, 1\% improvement, with a much shorter input sequence (156 vs 393). From the human evaluation, we notice a significant improvement in favor of GPT2 models, expecially GPT2+KE, in both humanness and correctness. 

\begin{table}[t]
\centering
\resizebox{0.48\textwidth}{!}{
\begin{tabular}{r|ccc|cc}
\hline
\textbf{Model} & \textbf{BLEU} & \textbf{F1} & \textbf{Succ.} & \textbf{Hum.} & \textbf{Corr.} \\ \hline
KB-Trs$^1$  &  14.80 & 45.30& - & - & - \\
MLMN$^2$  & 13.61 & 54.85& - & - & - \\ 
BoSsNet$^3$ & 15.20 & 43.10 & - & - & - \\
KBRet$^4$  & \textbf{18.64} & 55.76& 62.03 & 3.13 & 77.33 \\ 
\hline
GPT2  & 13.58 & 34.69& 30.38 & 3.42 & 66.67 \\
GPT2+KB  & 13.59 & 50.45& 62.03 & 2.42 & 70.37 \\
GPT2+KE  & 18.00 & \textbf{54.85}& \textbf{74.68} & \textbf{3.48} & \textbf{83.50} \\ \hline
Human  & - & -& 86.08 & 3.60 & 96.97 \\ \hline
\end{tabular}
}
\caption{Results on the CAMREST dataset. $^1$\cite{haihong2019kb}. $^2$\cite{reddy2019multi}.
$^3$\cite{raghu2019disentangling}. We re-evaluate $^4$\cite{qin2019entity} using our script that includes postcode as an entity and removes API-calls from F1-count.}
\label{tab:camrest-results}
\end{table}

\begin{table*}[t]
    \begin{minipage}{.62\linewidth}
        \centering
        \begingroup
            \resizebox{1.0\textwidth}{!}{
                \begin{tabular}{r|c|cccc|cc}
                \hline
                \textbf{Model} & \textbf{BLEU} & \textbf{Ent.} & \textbf{Nav.} & \textbf{Wea.} & \textbf{Sch.} & \textbf{Hum.} & \textbf{Cor.}  \\ \hline
                KVRet$^1$ & 13.20 & 48.00 & 44.50 & 53.30 & 62.90 & - & - \\ 
                MLMN$^2$ & 17.10 & 55.10 & 41.30 & 47.00 & 68.30 & - & - \\\hline
                BoSsNet$^3$ & 8.3 & 35.9 & - & - & - & - & - \\
                Mem2Seq$^4$ & 12.20 & 33.40 & 20.00 & 49.30 & 32.80 & - & - \\
                KBRet$^5$ & 13.90 & 53.70 & 54.50 & 52.20 & 55.60 & - & - \\
                KB-Trs$^6$ & 13.90 & 37.10 &  23.30 & 48.20 & 51.20 & - & - \\
                GLMP$^7$ & 13.90 & 60.70 & 54.60 & 56.50 & 72.50 & - & - \\
                DFF$^8$ & 14.40 & \textbf{62.70} & \textbf{57.90} & 57.60 & \textbf{73.10} & 3.28 & 68.90 \\ \hline
                GPT2 & 15.60 & 39.11 & 23.41 & 53.74 & 52.26 & \textbf{3.49} & 67.05 \\
                GPT2+KB & 17.03 & 58.60 & 48.37 & \textbf{62.87} & 72.22 & 3.47 & 81.03 \\
                GPT2+KE & \textbf{17.35} & \textbf{59.78} & \textbf{53.53} & 57.73 & \textbf{72.58} & 3.44 & \textbf{85.56} \\ \hline
                Human$^1$ & 13.50 & 60.70 & 55.20 & 61.60 &  64.30 & 3.54 &  97.92 \\ \hline
                \end{tabular}
            }
            \caption{Results on the SMD (KVR) dataset. $^{^1}$\citet{eric2017key} $^2$\cite{reddy2019multi}
            $^3$\cite{raghu2019disentangling} $^4$\cite{madotto2018mem2seq} $^5$\cite{qin2019entity} $^6$\cite{haihong2019kb} $^7$\cite{wu2019global} $^8$\cite{qin2020dynamic}} 
            \label{tab:SMD}
        \endgroup
    \end{minipage}%
    \hspace{8pt}
    \begin{minipage}{.35\linewidth}
        \centering
        \begingroup
        \includegraphics[width=\linewidth]{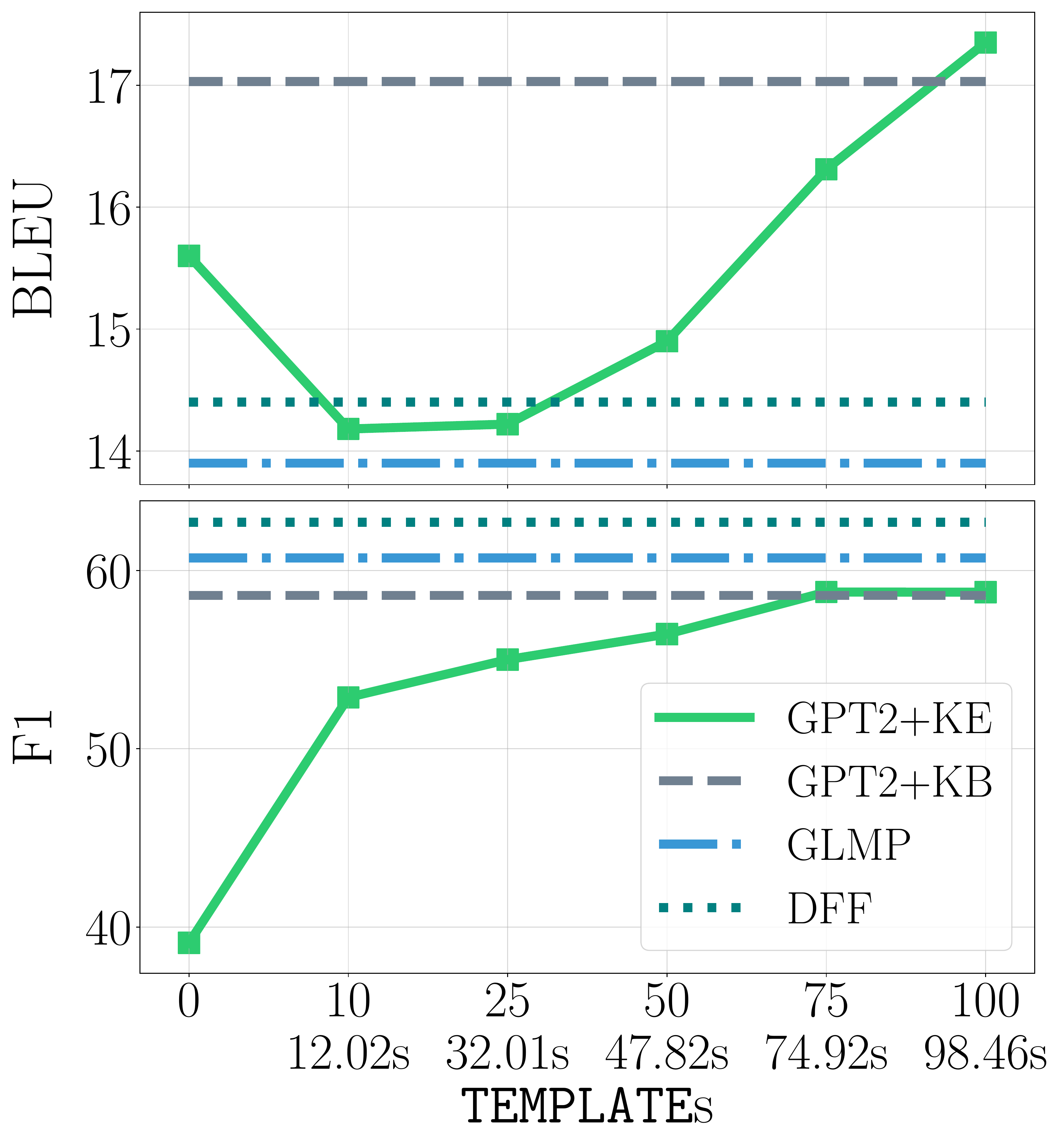}
        \vspace{-20pt}
        \captionof{figure}{BLEU and F1-Score versus number of \texttt{TEMPLATEs} in the SMD dataset.}
        \label{fig:SMD}
        \endgroup
    \end{minipage}%
\end{table*}
\paragraph{SMD} is a human-to-human collected dataset~\cite{ericKVR2017} with three domains: Navigation, Weather, and Calendar. In this dataset, no user goal query is provided; thus, we manually annotate 100 dialogues per domain from the training set, resulting in as many \texttt{TEMPLATES}. Moreover, to simplify the \texttt{KE-DELEX} function, we also tag the entities in the conversation. Differently from other datasets, the KB dynamically changes in each dialogue and thus requires a KB update operation. To cope with this setting, we propose a fine-tuning approach as follows: given a dialogue KB from the test set, 1) we use the \texttt{TEMPLATEs} and the corresponding user goal queries to generate the KE dialogues based on the KB, 2) we fine-tune the GPT2 model with the generated dialogues, and 3) we use the model to generate the response for the considered dialogue sample from the test set. Based on the KB size, for each test sample, we generate, on average, 469/162/6,629 KE dialogues for Navigate/Calendar/Weather, respectively.

Table~\ref{tab:SMD} compares the performance of our method with existing baselines. Firstly, we notice that GPT2, even without KB, performs better than the existing baselines~\cite{madotto2018mem2seq,haihong2019kb,raghu2019disentangling}, suggesting a significant overlapping between the training and test set KBs. As aforementioned, GPT2 with the KB as input does not perform as well as other baselines with a similar setting, except for the Weather domain, where it actually achieves SOTA performance. GPT2 fine-tuned with the KE dialogues performs almost as well as DFF~\cite{qin2020dynamic} in terms of F1-score, but from the human judgments, GPT2-based models perform significantly better both in terms of humanness and correctness. 

\begin{table*}[!t]
\resizebox{0.98\textwidth}{!}{
\begin{tabular}{r|cc|cc|ccccc|cc}
\hline
\textbf{Model}   &
\multicolumn{1}{c}{\textbf{Inform}} & \multicolumn{1}{c|}{\textbf{Success}} & \multicolumn{1}{c}{\textbf{BLEU}} & \multicolumn{1}{c|}{\textbf{F1}} & \multicolumn{1}{c}{\textbf{Train}} & \multicolumn{1}{c}{\textbf{Attraction}} & \multicolumn{1}{c}{\textbf{Hotel}} & \multicolumn{1}{c}{\textbf{Rest}} & \multicolumn{1}{c|}{\textbf{Taxi}} & \multicolumn{1}{c}{\textbf{Human}} & \multicolumn{1}{c}{\textbf{Correct}} \\ \hline
Mem2Seq$^1$ & - & - & 6.60 & 21.62 & - & 22.00 & 21.00 & 22.40 & - & - & - \\
DSR$^2$ & - & - & 9.10 & 30.00 & - & 28.00 & 27.00 & 33.40 & - & - & - \\
GLMP$^3$ & - & - & 6.90 & 32.40 & - & 24.40 & 28.10 & 38.40 & - & - & - \\
DFF$^4$ & - & - & 9.40 & 35.10 & -  & 28.10 & 30.60 & \textbf{40.90} & - & 2.65 & 25.53 \\ \hline
GPT2 & 64.60 & 51.77 & 14.33 & 30.38 & 23.30  &  15.11  &  23.56  &  25.62  & 89.76 & 3.51 & 55.91 \\
GPT2+KE & \textbf{72.57}  &  \textbf{64.16}  & \textbf{15.05}  &  \textbf{39.58} &  \textbf{23.79}  &  \textbf{43.32}  &  \textbf{33.44}  &  37.10  & \textbf{92.38} & \textbf{3.56} & \textbf{73.38} \\ \hline
DAMD$^{\star}$ & 72.12  &  61.06 & 11.48 & - & - & - & - & - & - & 3.31 & 67.97 \\ \hline
Human & - & - & - &  - & - & - & - & - & - & 3.66 & 96.85 \\ \hline
\end{tabular}
}
\caption{Results on the MultiWOZ dataset. $^1$\cite{madotto2018mem2seq}. $^2$\cite{wen2018sequence}. $^3$\cite{wu2019global}. $^4$\cite{qin2020dynamic}. $^{\star}$We evaluate DAMD~\cite{zhang2019task} with our plain text scorer.}
\label{tab:MWOZ}
\end{table*}

\paragraph{MultiWOZ} dataset~\cite{budzianowski2018multiwoz} consists of five domains: Train, Attraction, Hotel, Restaurant, and Taxi. Following \citet{qin2020dynamic}, we select only the dialogues with a single domain, which is more challenging since less data is available, and we leave the multiple domains per dialogue to future work. This dataset provides both the user goal query and the span annotation for the entities. The \texttt{KE-DELEX} function is implemented using the entity span annotation, although advanced string matching could also work. We extracted 63/116/289/59 \texttt{TEMPLATEs} and 3,826/2,495/21,970/30,149 KE dialogues for Attraction/Hotel/Restaurant/Train, respectively. The Taxi domain does not have a KB, since all of its dialogues are booking related.

In Table~\ref{tab:MWOZ} we compare GPT2 trained with KE dialogues with the current state-of-the-art for pipelined models (DAMD)~\cite{zhang2019task} and end-to-end models (DFF)~\cite{qin2020dynamic}. We re-train DAMD on single domain dialogues, and we use the script provided by the authors to relexicalize the generated templates. We are aware of newly-released models~\cite{hosseini2020simple,peng2020soloist}; however, no code was available at submission time for running the results on single domain. 

In DFF, we used the provided model to generate the system responses for the human evaluation, but we could not use our scorer to automatically evaluate the Inform, Success, and F1 since no dialogue Id was present in their pre-processed data.\footnote{We reproduce their generated responses from https://github.com/LooperXX/DF-Net} Moreover, the authors provided the results in three domains (Attraction, Hotel, Restaurants) for multiple baselines by using the Gold-KB as input. 


From our experiments, two points can be highlighted: 1) GPT trained with KE dialogues performs as well as DAMD trained using DST and template responses, in both automatic and human evaluation. Using the original scorer~\cite{budzianowski2018multiwoz}, DAMD achieved 85.40 Inform and 70.40 Success score, but when the responses are relexicalize and we use our scorer, the results are significantly lower.\footnote{We properly align the entities to our scorer.} The human evaluation confirms the correctness of our plain scorer and it shows that the relexicalization process is not a trivial task; 2) Our model achieves a higher BLEU and F1-score that other models trained with gold KB as input, and it achieve a significantly higher correctness compare to DFF. This is easily explainable by the fact that DFF does not issue booking API and thus it constantly mistakes the booking results. In appendix H, we show how our model handles the booking API.  

\begin{table}[t]
\centering
    \resizebox{0.45\textwidth}{!}{
        \begin{tabular}{r|c|ccc}
        \hline
        \multicolumn{1}{c|}{\textbf{Model}} & \textbf{Iter.} & \textbf{BLEU} & \textbf{Prec.} & \small{\textbf{\begin{tabular}[c]{@{}c@{}}OOV \\ Prec.\end{tabular}}} \\ \hline
        
        GPT2+PATH & - & \textbf{7.32} & \textbf{86.41} & \textbf{5.55} \\ \hline
        GPT2 & - & 4.89 & 76.85 & 0.66 \\
        GPT2+KE & 3K & \textbf{\textit{5.04}} & 79.14 & 1.01  \\
        GPT2+KE & 6K & 5.00 & 78.87 & 1.40  \\
        GPT2+KE & 9K & 4.72 & \textbf{\textit{79.41}} & 1.65 \\
        GPT2+KE & 12K & 4.64 & 78.59 & \textbf{\textit{2.11}} \\ \hline
        \end{tabular}
    }
    \caption{Results on the OpenDialKG dataset. PATH represents the model with the correct nodes and relations provided from the dataset.} \label{DIALKG-RESULTS}
\end{table}

\paragraph{OpenDialKG} is a human-to-human collected dataset~\cite{moon2019opendialkg} consisting of four domains: Music, Sport, Book, and Movie. No official split is provided and thus we randomly split the dataset in 80/10/10 for the train/valid/test, respectively. The dataset provides a large knowledge graph with 100K entities and 1.1M relations, and the annotated entity path that connects $U_t$ and $S_t$. The graph relations in the annotated path are the user goal query defined in Equation~\ref{KB_PATH}, but after a careful analysis, we discover that the annotation is incomplete in most of the dialogues. Therefore, we decided to automatically generate the user goal queries using string matching and the \texttt{CYPHER} query language.\footnote{More details in Appendix A.1} This process generates 11K possible \texttt{TEMPLATEs}, which, if used over the user goal query output, generate over a billion KE dialogues. This is because the knowledge graph is large, and each user goal query returns a large number of equivalent entities. To overcome this issue, 1) we select a subset of the knowledge graph, 5,691 entities, and 39,728 relations, which covers most of the test set entities, and 2) we iteratively generate dialogues by sampling \texttt{TEMPLATES} and using \texttt{KE-RELEX} over the sampled query results.

Table~\ref{DIALKG-RESULTS} compares a GPT2 trained with the provided gold path as input with a GPT2 trained on an increasing number of dialogues generated by the iterative procedure. We observe that by increasing the number of iterations, thus the number of KE dialogues, the entity F1-score increases, especially for OOV entities, but at the same time, the BLEU score decreases. After a careful qualitative analysis, we notice that the string matching algorithm used for extracting the user goal queries generate noisy and incomplete \texttt{TEMPLATE}s, and thus most of the KE dialogues have imprecise knowledge. We leave the annotation of the user goal queries and the human evaluation to the future work. 

\section{Analysis and Discussions}

\paragraph{Templates vs. Performance} In all experiments, we show that given the generated KE dialogues, the model learns to embed the KB into its parameters. However, the user goal query still requires human annotations; thus, we want to analyze the effect of using increasingly less \texttt{TEMPLATE}s in KE. For instance, in Figure~\ref{fig:SMD}, we report the number of \texttt{TEMPLATE}s used for fine-tuning versus the BLEU score and the entity F1-score in the SMD dataset. In general, we observe that more \texttt{TEMPLATE}s increase significantly both the F1 and BLEU score. Especially, we observe that BLUE score linearly increase with the number of \texttt{TEMPLATE}s used in training, suggesting that a more diverse and fluent generation can be achieved using more \texttt{TEMPLATE}s. In Appendix F, we report the same analysis in each datasets, where we observe a similar trend. 

\paragraph{Limitation \& Dynamic KB}
Throughout our experiments, we identify two major limitations: noisy KE dialogues generation and fine-tuning time for dynamic KBs. Although the proposed KE results successfully embed the KB into the model parameters, the generated KE dialogues are sometimes noisy. For example, the \texttt{KE-DELEX} function converts, ``i want to find an expensive restaurant..." into a \texttt{TEMPLATE} ``i want to find an [price-range] restaurant...". Then the \texttt{KE-RELEX} can generate ``i want to find a cheap restaurant...", which has a clear grammar mistake. This type of error does not happen often, and we notice that GPT2 is robust to this kind of noisy input. In future work, we propose to improve the robustness and fluency of our model using different regularization losses. Moreover, in the case of dynamic KBs a substantial fine-tuning cost is required for updating the KB. Figure~\ref{fig:SMD} shows the average time-per-epoch spent for fine-tuning in SMD. In future work, we propose to study both a meta-learning~\cite{finn2017model} strategy for quick fine-tuning and continual learning approach for updating the KB while retaining the previous existing knowledge.

\section{Related Work}
\paragraph{Dialogue Systems} are categorized~\cite{gao2018neural} into chit-chat~\cite{vinyals2015neural,serban2016generative} and task-oriented~\cite{williams2007partially,young2013pomdp}; in this paper we focus on the latter. Task-oriented dialogue systems are further classified into: modularized~\cite{levin2000stochastic,hori2009statistical,lee2009example}, retrieval~\cite{henderson2019training,wu2020tod} end-to-end~\cite{bordes2016learning,ericKVR2017,eric-manning:2017:EACLshort,reddy2019multi,madotto2018mem2seq,wu2019global,madotto2020attention,neelakantan2019neural,qin2019entity,qin2020dynamic,raghu2019disentangling,haihong2019kb,9053667} and hybrid~\cite{shu2018incorporating,lei2018sequicity,zhang2019task,mehri2019structured,chen2019semantically,peng2020soloist,ham-etal-2020-end,hosseini2020simple,le2020uniconv,lin2020mintl}. 
To the best of our knowledge, these methods use either DST/S-ACT annotations, template responses, or all/partial KB as the input to the model, where instead we only use the dialogue history. 

Recently, several task-oriented dialogue models are introduced to tackle the resource scarcity challenges in target domains~\cite{bapna2017towards,shah2019robust,wu2019transferable,liu2020coach} and target languages~\cite{mrkvsic2017semantic,schuster2019cross,chen2018xl,liu2019zero}, and large pre-trained language models are shown to possess the capability to quickly adapt to task-oriented dialogue tasks by using only a few data samples~\cite{peng2020few,madotto2020language,wu2020tod}.

\paragraph{Data Augmentation} is a widely used technique to improve both robustness and performance~\cite{guo2019augmenting,yang2020g}. Task-oriented dialogue systems have been explored to improve DST~\cite{song2020data,yoo2020variational,campagna2020zero}, Natural Language Understanding (NLU)~\cite{peng2020data}, intent classification~\cite{kumar2019closer} and hybrid end-to-end systems~\cite{zhang2019task,rastogi2019scaling}. These data augmentation methods aim to improve the final performance of the given task, e.g., zero-shot performance, template response, etc., where instead, our proposed approach aims to store the KB into the model parameters. 

\paragraph{Agenda-Based User Simulation} builds an interactive system that models the user turns~\cite{schatzmann2007agenda} rather than the system. User simulators are designed to cover all possible user queries while keeping a diverse and fluent user interaction. This enables models to learn a better dialogue policy via interaction~\cite{asri2016sequence,li2017end,wu2019switch,peng2018deep}, and it is especially useful in scenarios in where few or no data is available~\cite{liu2017iterative,liu2017end,shah2018bootstrapping,kreyssig2018neural,li2020guided}. In our work, instead, we use all the possible user goal queries to generate dialogues directly, instead of creating a reinforcement learning loop to train the model. 

\paragraph{Language Models as Knowledge Bases} has been used for encoding common sense knowledge into transformers~\cite{bosselut2019comet,liu2019k,xiong2019pretrained,wang2020k,wang2019kepler}. \cite{guan2020knowledge} improved story generation by training a Language Model with knowledge triples converted into sentences using predefined templates~\cite{levy2017zero}. Differently, we extract templates from real data, and we aim to store the KB into the models parameters to be able to extract knowledge directly, instead of improving common sense generation. Moreover, several studies tried to extract~\cite{petroni2019language,kassner2019negated,petroni2020context} or use~\cite{roberts2020much} large pre-trained models, e.g. BERT~\cite{devlin2019bert}, as knowledge bases. 

\section{Conclusion}
In this paper, we propose to learn the KB directly into the model parameters using a novel Knowledge Embedded approach, that is fundamentally different from giving the KB as input or using the DST for querying the KB. We demonstrate that our approach is scalable to different KB sizes and it can be used with dynamically changing KBs via fine-tuning. Automatic and human evaluations confirm that models with embedded KBs achieve competitive performance in all evaluated datasets. Finally we show, for the first time, that end-to-end models can perform as well as pipelined modularized systems~\cite{zhang2019task} in the MWoZ single domain dataset. 

\section{Acknowledgements}

This work has been partially funded by MRP/055/18 of the Innovation Technology Commission, The Hong Kong SAR Government. 

\bibliography{anthology,emnlp2020}
\bibliographystyle{acl_natbib}

\appendix

\setcounter{table}{0} 
\setcounter{figure}{0}
\renewcommand{\thetable}{\Alph{section}\arabic{table}}
\renewcommand\thefigure{\Alph{section}\arabic{figure}} 

\section{Knowledge Embedded}
\label{sec:appendix_A}

We provide intuitive samples of our Knowledge Embedded approach in different datasets. Table \ref{tab:delex_SMD} and Table \ref{tab:delex_BABI} shows the user goal query in form of \texttt{SQL} syntax for tabular-formatted KB and how the \texttt{KE-DELEX} generate \texttt{TEMPLATE}s. Similarly Table~\ref{tab:delex_OPENDIALKG} shows the user goal query in \texttt{CYPHER} syntax for graph-formatted KB and how the \texttt{KE-DELEX} generates \texttt{TEMPLATE}s. We further discuss the detail of the \texttt{KE-DELEX} for OpenDialKG in the following section.

\subsection{OpenDialKG Knowledge Embedded}
In OpenDialKG, we divide the \texttt{KE-DELEX} process into three steps: string matching, spanning tree, and dialogue generation. We perform string matching using cased letters, and we only select the entities with a minimum length of five characters to reduce the detection of false entities. To handle overlapping sequences, such as ``The Dark" and ``The Dark Knight" in ``I enjoy watching The Dark Knight", we perform a further filtering in each turn and we take the longest string when there is an overlapping between two or more entities.

\paragraph{String Matching Process} We extract a set of entities that from in the dialogue based on the nodes in the graph. This set of entities are defined as the $R$ of a user goal. To complete the user goal, we need to find the constraint $C$. This can be done by generating a spanning tree from the Knowledge Graph between all entities in $R$.

\paragraph{Spanning Tree} We get all the relations and intermediary nodes between each pair of nodes in $R$. The collected relations are what we defined as constraint $C$ of the user goal. With the given $R$ and $C$, we can build a \texttt{CYPHER} query in form of $\texttt{MATCH} \ C \ \texttt{RETURN}\ R$ as mentioned in the Methodology. 

\begin{figure}[t]
    \centering
    \includegraphics[scale=0.5]{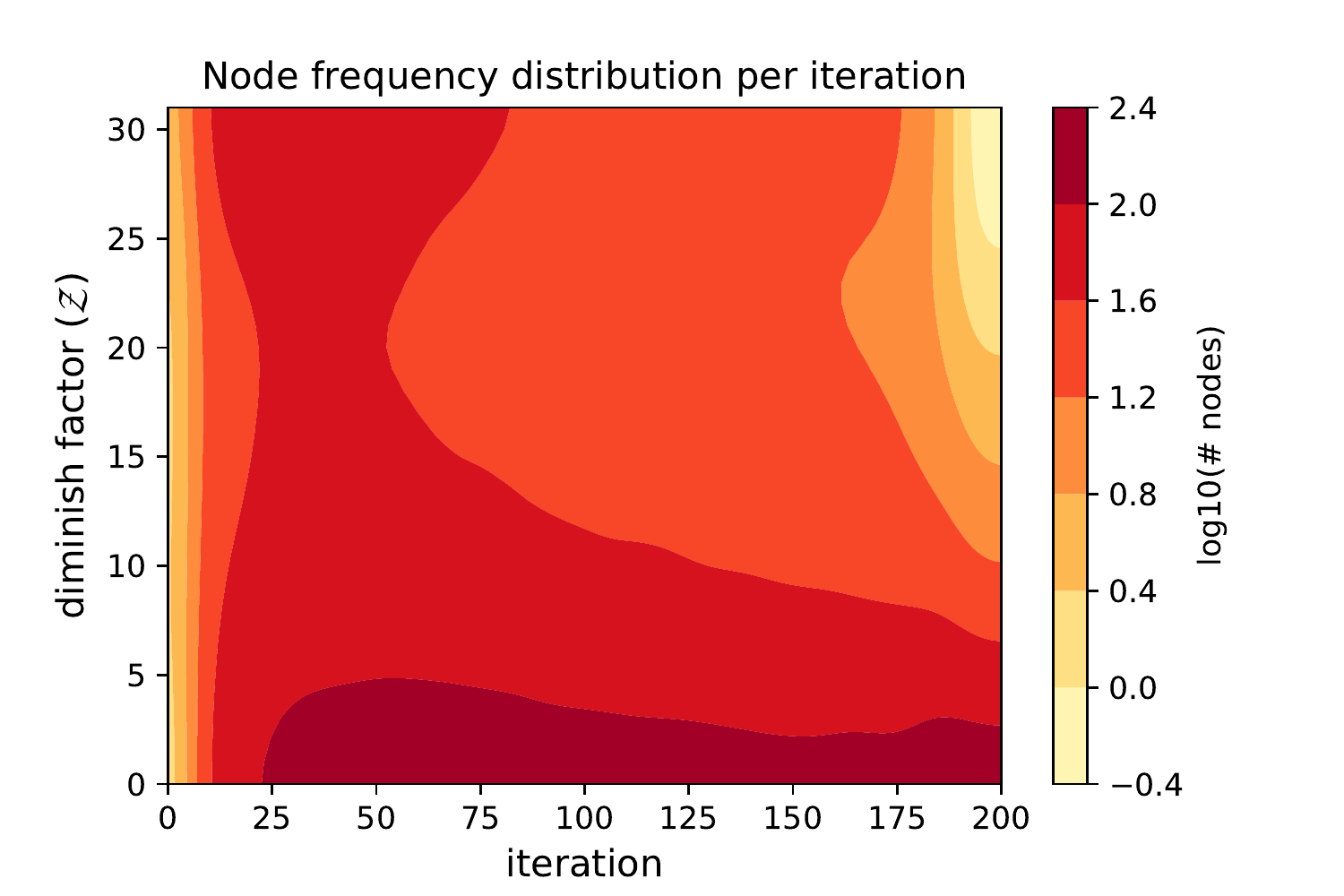}
    \caption{Distribution of \#nodes over $\mathcal{Z}$ and iteration.}
    \label{fig:sampling-distribution}
\end{figure}

\paragraph{Dialogue Generation} We use the \texttt{CYPHER} query to retrieve the equivalent nodes for the dialogue using neo4j, a graph database which supports diverse functionality for graph retrieval and manipulation. An example of our query generation is shown in Table \ref{tab:delex_OPENDIALKG}. To ensure diversity of the dialogue generation, we set up a diminishing factor $\mathcal{Z}$ on each node, to restrict the access to the same node over time. We initialize $\mathcal{Z}$ with the number of edges on each node, and we decremented $\mathcal{Z}$ each time the node is used for the generation. In order to constraint the query with the limiting factor $\mathcal{Z}$, we expand the \texttt{CYPHER} query into $\texttt{MATCH} \ C \ \texttt{WHERE} \ \mathcal{Z}_n > 0 \ \forall \ n \in \{C,R\} \ \texttt{RETURN}\ R$. We iteratively generate dialogues by sampling \texttt{TEMPLATE}s. For each iteration, we randomly sampled 200 \texttt{TEMPLATE}s and use \texttt{KE-RELEX} to generate the dialogues. To check the diversity of the entity in the generated dialogues, we measure the number of nodes per $\mathcal{Z}$ per iteration. As shown in Figure~\ref{fig:sampling-distribution}, the nodes with high $\mathcal{Z}$ is reduced over iteration and on each iteration, more and more nodes reach $\mathcal{Z} = 0$, which ensure that the entity selected for the generation of the same \texttt{TEMPLATE} would include a different set of entities.

\begin{table*}[t]
    \begin{minipage}{.49\linewidth}
    \centering
        \begingroup
        \setlength{\tabcolsep}{3.3pt} 
        \begin{tabular}{rl}
        \multicolumn{2}{c}{SMD} \\ \hline\hline
        \textbf{U:} & Where is the closest \hlblue{gas station}? \\
        \textbf{S:} & \hlgreen{Valero} is \hlorange{3 miles} away. \\
        \textbf{U:} & What is the address? \\
        \textbf{S:} & \hlgreen{Valero} is located at \hlred{200 Alester Avenue}. \\ \hline\hline
        \multicolumn{2}{l}{\begin{tabular}[c]{@{}l@{}}\textbf{SELECT} \hlblue{type}, \hlgreen{poi}, \hlorange{dist}, \hlred{addr}\\ \textbf{FROM} navigation\\ \textbf{GROUP BY} \hlblue{type}\\ \textbf{HAVING} \hlorange{dist} = \textbf{MIN}(\hlorange{dist})\end{tabular}}\\ \hline
        \end{tabular}
        \caption{Example of user goal query from SMD~\cite{ericKVR2017}. The requested information $R$ are the attribute \textit{type}, \textit{poi}, \textit{dist}, and \textit{addr}, with the constrain $C$ using as \texttt{OP}: \texttt{GROUP BY}, \texttt{HAVING} and \texttt{MIN}. The dialogues is filled with an instance from the query results, with the entities marked with the corresponding colour.}
        \label{tab:delex_SMD}
        
        \begin{tabular}{rl}
        \multicolumn{2}{c}{bAbI5} \\ \hline \hline
        \textbf{U:} & \begin{tabular}[c]{@{}l@{}}May I have a table for \hlyellow{two} people \\ in a \hlblue{cheap} price range?\end{tabular} \\
        \textbf{S:} & Any preference for a type of cuisine? \\
        \textbf{U:} & With British cuisine. \\
        \textbf{S:} & Where should it be? \\
        \textbf{U:} & In \hlgreen{Rome}, thanks. \\
        \textbf{S:} & Let me look for some options \\
        \textbf{U:} & Instead could it be with \hlorange{Italian} food? \\
        \textbf{S:} & Sure, is there anything else? \\
        \textbf{U:} & No, thanks. \\
        \textbf{S:} & How about \hlred{Pizza Hut}? \\ \hline
        \hline
        \multicolumn{2}{l}{\begin{tabular}[c]{@{}l@{}}\textbf{SELECT} \hlorange{cuisine}, \hlgreen{loc}, \hlyellow{ppl}, \hlblue{price}, \hlred{name}\\\textbf{FROM} restaurant\end{tabular}}\\ \hline
        \end{tabular}
        \caption{Example of user goal query in SQL formation. The user goal specify a cuisine (i.e., Italian), but in the dialogue the user mention multiple once. To resolve this tie, we select the last mentioned cuisine entities in the dialogue.   }
        \label{tab:delex_BABI}
        
        \endgroup
    \end{minipage}%
    \hfill
    \begin{minipage}{.49\linewidth}`
        \setlength{\tabcolsep}{3.3pt} 
        \begin{tabular}{rl}
        \multicolumn{2}{c}{OpenDialKG} \\ \hline\hline
        \textbf{U:} & \begin{tabular}[c]{@{}l@{}}Any movies similar to \hlblue{Gangs of New York} \\ that you can recommend?\end{tabular} \\
        \textbf{S:} & \begin{tabular}[c]{@{}l@{}}Sure, \hlgreen{Quantum of Solace} has the same actor \\ \hlred{Tim Pigott-Smith}.\end{tabular} \\
        \textbf{U:} & Is that the one with \hlorange{Daniel Craig}? \\
        \textbf{S:} & Yes, it is a \hlyellow{thriller} also starred by \hlorange{Daniel Craig}. \\
        \textbf{U:} & I really love \hlyellow{thriller}s. Any suggestion? \\
        \textbf{S:} & \begin{tabular}[c]{@{}l@{}}\hlorange{Daniel Craig} also starred in \hlpurple{The Girl} \\ \hlpurple{with the Dragon Tattoo}\end{tabular} \\
        \textbf{U:} & Thanks for the suggestion \\ \hline\hline
        \multicolumn{2}{l}{
        \quad\begin{minipage}{0.46\linewidth}
        \begin{tabular}[c]{@{}l@{}}\textbf{MATCH}\\ \hlred{n1}-[ActorsIn]$\rightarrow$ \hlblue{n2},\\ \hlred{n1}-[ActorsIn]$\rightarrow$ \hlgreen{n3},\\ \hlorange{n4}-[ActorsIn]$\rightarrow$ \hlgreen{n3},\\ \hlorange{n4}-[ActorsIn]$\rightarrow$ \hlpurple{n6},\\ \hlgreen{n3}-[HasGenre]$\rightarrow$ \hlyellow{n5},\\ \hlpurple{n6}-[HasGenre]$\rightarrow$ \hlyellow{n5}\\ \textbf{RETURN} \hlred{n1}, \hlblue{n2}, \hlgreen{n3},\\ \hlorange{n4}, \hlyellow{n5}, \hlpurple{n6}\end{tabular}
        \end{minipage}
        \begin{minipage}{0.4\linewidth}
        \centering
        \end{minipage}
        \hfill 
        } \\ \hline
        \end{tabular}
        \caption{Example of user goal query from OpenDialKG \cite{moon2019opendialkg} with \texttt{CYPHER}  syntax ~\cite{webber2012programmatic}, where the nodes are the requested information in $R$, and the labeled edges the constrains in $C$.}
        \label{tab:delex_OPENDIALKG}
    \end{minipage}%
\end{table*}

\begin{table*}[t]
\centering
\resizebox{0.98\textwidth}{!}{
\begin{tabular}{rcccccccc|l}
\multicolumn{1}{c}{\textbf{}} & \multicolumn{3}{c|}{\textbf{Pre-Processing}} & \multicolumn{5}{c|}{\textbf{Training/Testing}} & \multirow{2}{*}{\textbf{Model}} \\ \cline{1-9}
\multicolumn{1}{c|}{\textbf{}} & \textit{\textbf{Goal}} & \textit{\textbf{Span}} & \multicolumn{1}{c|}{\textit{\textbf{KB}}} & \textit{\textbf{DST}} & \textit{\textbf{S-ACT}} & \textit{\textbf{KB}} & \textit{\textbf{API}} & \textit{\textbf{LEX-R}} &  \\ \hline
\multicolumn{1}{r|}{\textit{E2E+Pipelined}} & \xmark & \cmark & \multicolumn{1}{c|}{\xmark} & \cmark & \cmark/\xmark & \cmark & \xmark & \cmark & \begin{tabular}[c]{@{}l@{}}Sequicity~\cite{lei2018sequicity}, DAMD~\cite{zhang2019task},\\Structured Fusion~\cite{mehri2019structured}, HDSA~\cite{chen2019semantically},\\ UniConv~\cite{le2020uniconv}, Soloist~\cite{peng2020soloist}, \\SimpleTOD~\cite{hosseini2020simple},\\MultiWOZ Benchmark~\cite{budzianowski2018multiwoz}\end{tabular} \\ \hline
\multicolumn{1}{r|}{\textit{E2E+API+KB}} & \cmark & \xmark & \multicolumn{1}{c|}{\cmark} & \cmark & \xmark & \cmark & \cmark & \xmark & \begin{tabular}[c]{@{}l@{}}MemoryNet~\cite{bordes2016learning}, \\Copy-Augmented Seq2Seq~\cite{eric-manning:2017:EACLshort},\\ Mem2Seq~\cite{madotto2018mem2seq}, MLMN~\cite{reddy2019multi},\\ GLMP~\cite{wu2019global}, BoSsNet~\cite{raghu2019disentangling},\\ KB-Trs~\cite{haihong2019kb}\end{tabular} \\ \hline
\multicolumn{1}{r|}{\textit{E2E+GOLD KB}} & \cmark & \xmark & \multicolumn{1}{c|}{\cmark} & \xmark & \xmark & \cmark & \xmark & \xmark & \begin{tabular}[c]{@{}l@{}} KVRet~\cite{ericKVR2017}, Mem2Seq~\cite{madotto2018mem2seq},\\ KBRet~\cite{qin2019entity},\\ Neural Assistant~\cite{neelakantan2019neural}, GLMP~\cite{wu2019global},\\ DFF~\cite{qin2020dynamic}, GCN~\cite{banerjee2019graph},\end{tabular} \\ \hline
\multicolumn{1}{r|}{\textit{E2E+KB}} & \xmark & \xmark & \multicolumn{1}{c|}{\xmark} & \xmark & \xmark & \cmark & \xmark & \xmark & Neural Assistant~\cite{neelakantan2019neural} \\ \hline
\multicolumn{1}{r|}{\textit{OURS}} & \cmark & \cmark & \multicolumn{1}{c|}{\cmark} & \xmark & \xmark & \xmark & \xmark & \xmark & KE-Dialogue \\ \hline
\end{tabular}
}
\caption{Comparison between different task-oriented methodologies in terms of annotation and mechanism used during pre-processing, training, and inference. \textit{Goal} denotes user goal, \textit{Span} denotes dialogue span, \textit{KB} denotes knowledge base , \textit{DST} denotes dialogue state tracking, \textit{S-ACT} denotes speech act, \textit{API} denotes  API call, and \textit{LEX-R} denotes lexicalization for the responses.}
\label{comparison}
\end{table*}

\section{Human Evaluation}
In this section, we show the annotators instructions used the for the human evaluation.

\subsection{Instructions for Humanness Evaluation}
\paragraph{Overview} In this task, you will be given a dialogue and a response, and you have to provide a rating of the response from 1 to 4 to indicate how human-like is the response. For instance, 4 means that the response is a very natural human response, and 1 indicates the response is obviously not a human-generated response.

\paragraph{Steps}
The steps of the humanness evaluation are as following:
\begin{itemize}
    \item There is a pre-filled columns with the dialogue history and a second column filled with the response text. 
    \item There is 1 blank humanness column where you can put rating from 1 to 4, indicating how human-like is the response: 4 indicates the response is a very natural human response and 1 indicates the response is obviously not a human-generated response.
    \item 1. Read the dialogue from the first column.
    \item 2. Read the response from the second column.
    \item 3. Rate how human-like is the response and fill the humanness rating on the third column.
\end{itemize}

\subsection{Instructions for Correctness Evaluation}
\paragraph{Overview} In this task, you will be given a KB, a dialogue history, and a response, and you have to provide a number of entity appearing in the KB and present in the response. You then need to check whether each of the entity is correct given the dialogue history, and the provided KB.

\paragraph{Steps}
The steps of the correctness evaluation are as following:
\begin{itemize}
   \item There are 3 pre-filled columns, the first column is the ID to the KB, if the KB is dynamic else -1, the second column contains the dialogue history of the conversation, and the third column contains the response. 
    \item There is 2 blank column, the first column (num\_entity) is where you can put the number of entities existing in the response text and second column (correct\_entity) is where you can put the number of correct entities based on the dialogue history and the KB.
    \item Another file for the KB is also provided in separate file named KB.txt
    \item 1. Read the dialogue history and the response from the second and third column.
    \item 2. Count how many entities on the response text that appears in the KB.
    \item 3. Find all the possible entities in the KB from the given the response on dialogue history and response and fill the num\_entity column.
    \item 4. Decide whether the entities in the response are in one of the possible entities in the KB.
    \item 5. Check whether the entities in the response text answer the given dialogue history or not (you need to make sure that the relation between each entity's attribute are also correct)
    \item 6. Count the number of correct entities attributes in the given text and fill the correct\_entity column
\end{itemize}

\subsection{Human Evaluation Results}
\begin{figure*}[t]
    \begin{minipage}{.49\linewidth}
        \centering
        \begingroup
            \includegraphics[width=\linewidth]{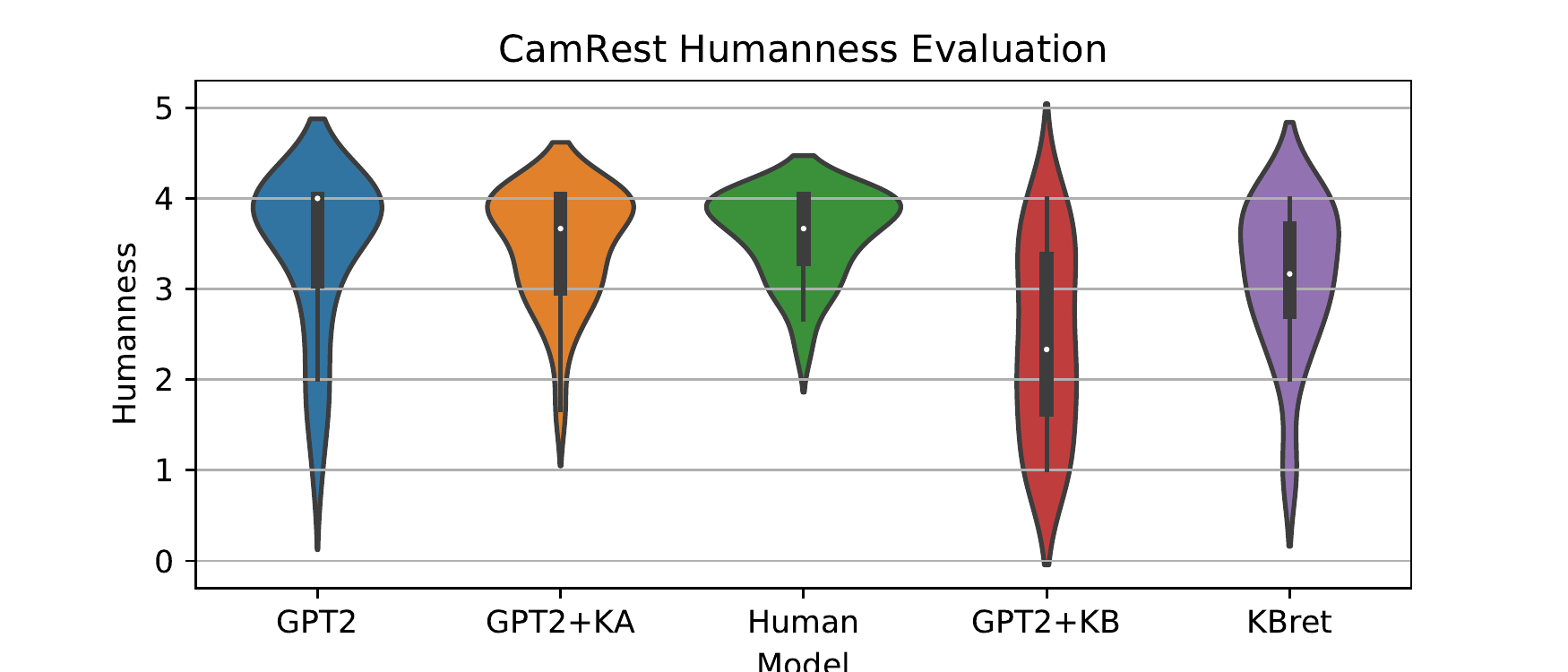}
            \noindent
            \includegraphics[width=\linewidth]{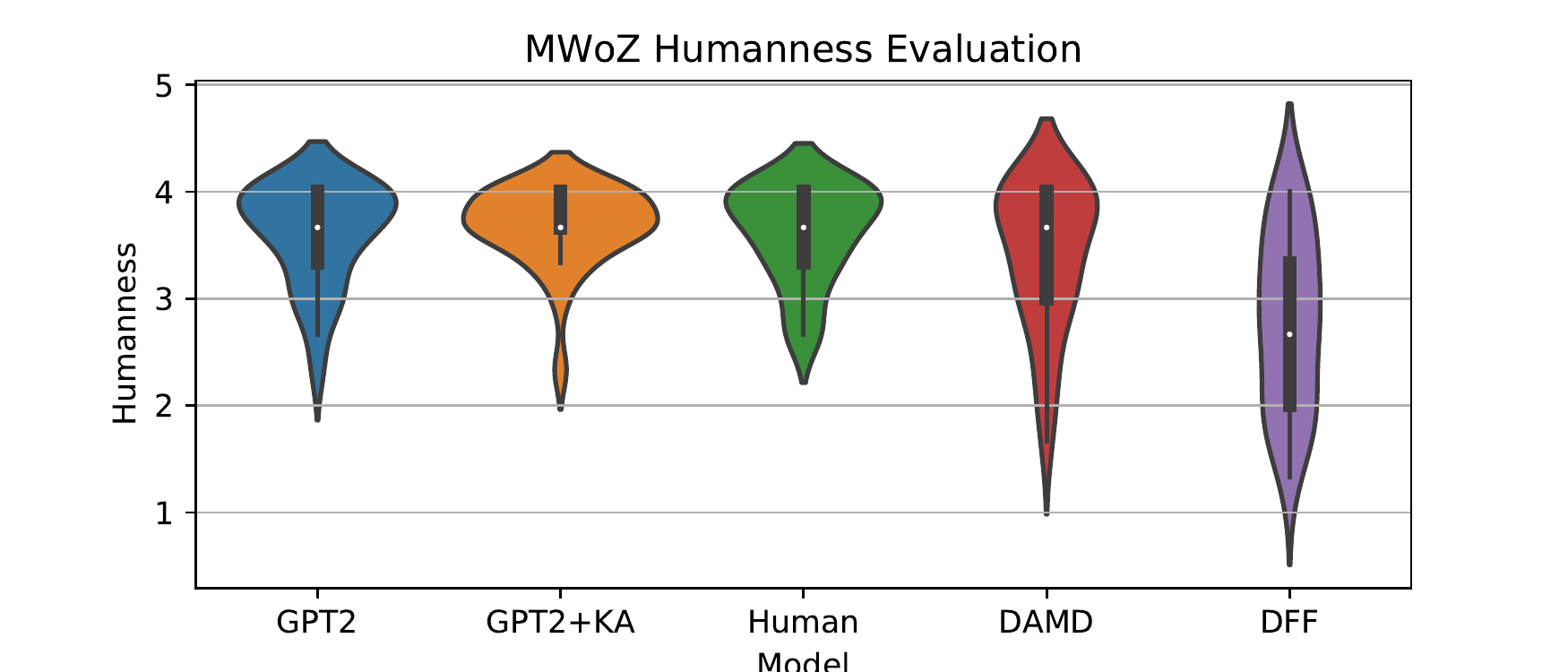}
            \noindent
            \includegraphics[width=\linewidth]{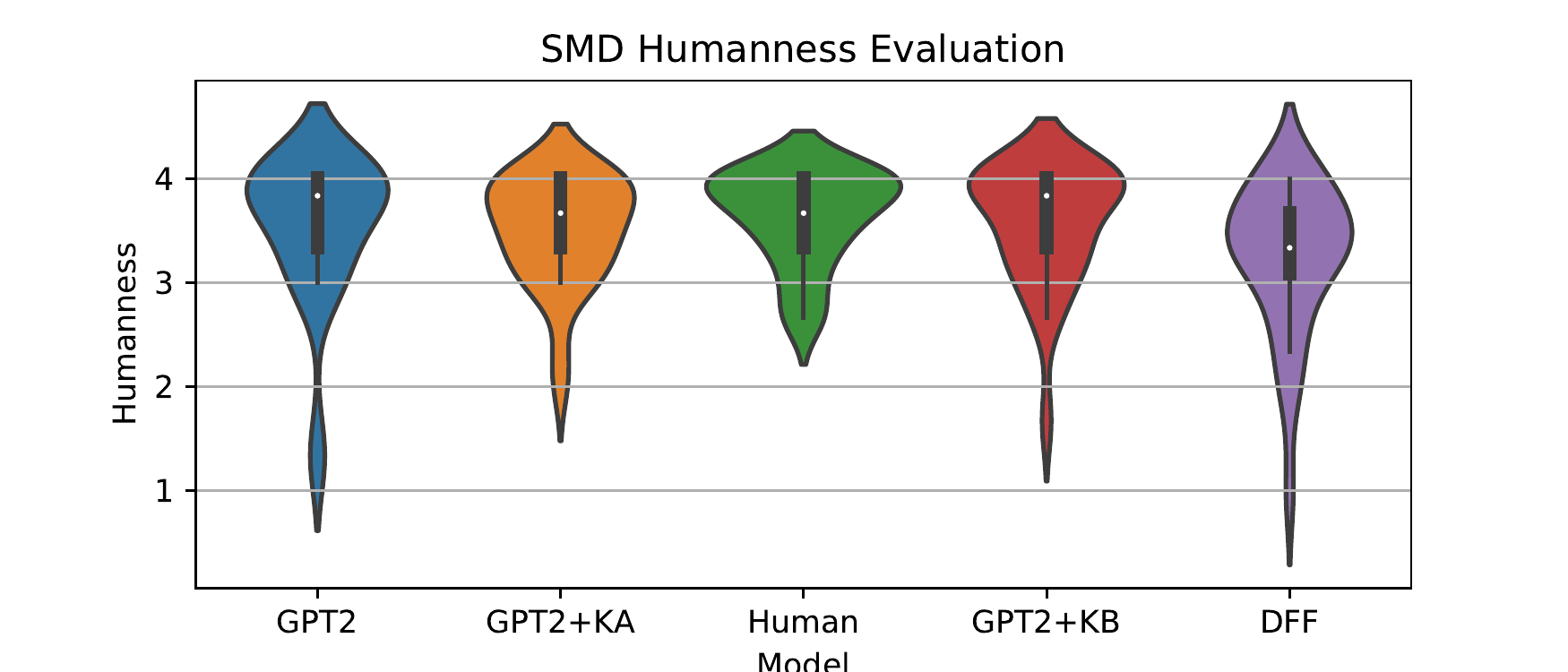}
            \caption{Humanness evaluation in CamRest, MWoZ, and SMD dataset.}
            \label{fig:humanness}
        \endgroup
    \end{minipage}
    \hfill
    \begin{minipage}{.49\linewidth}
        \centering
        \begingroup
            \includegraphics[width=0.9\linewidth]{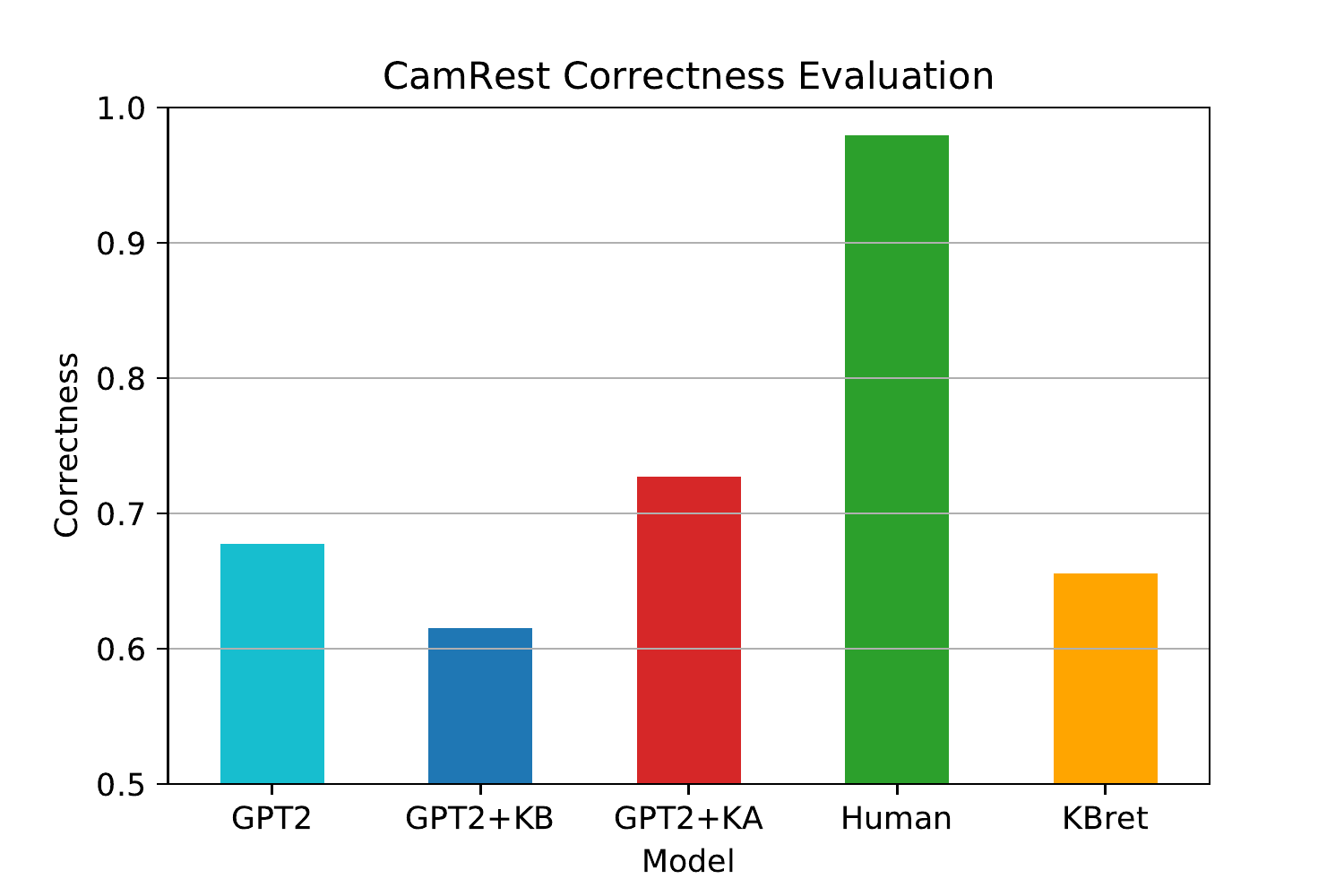}
            \noindent
            \includegraphics[width=0.9\linewidth]{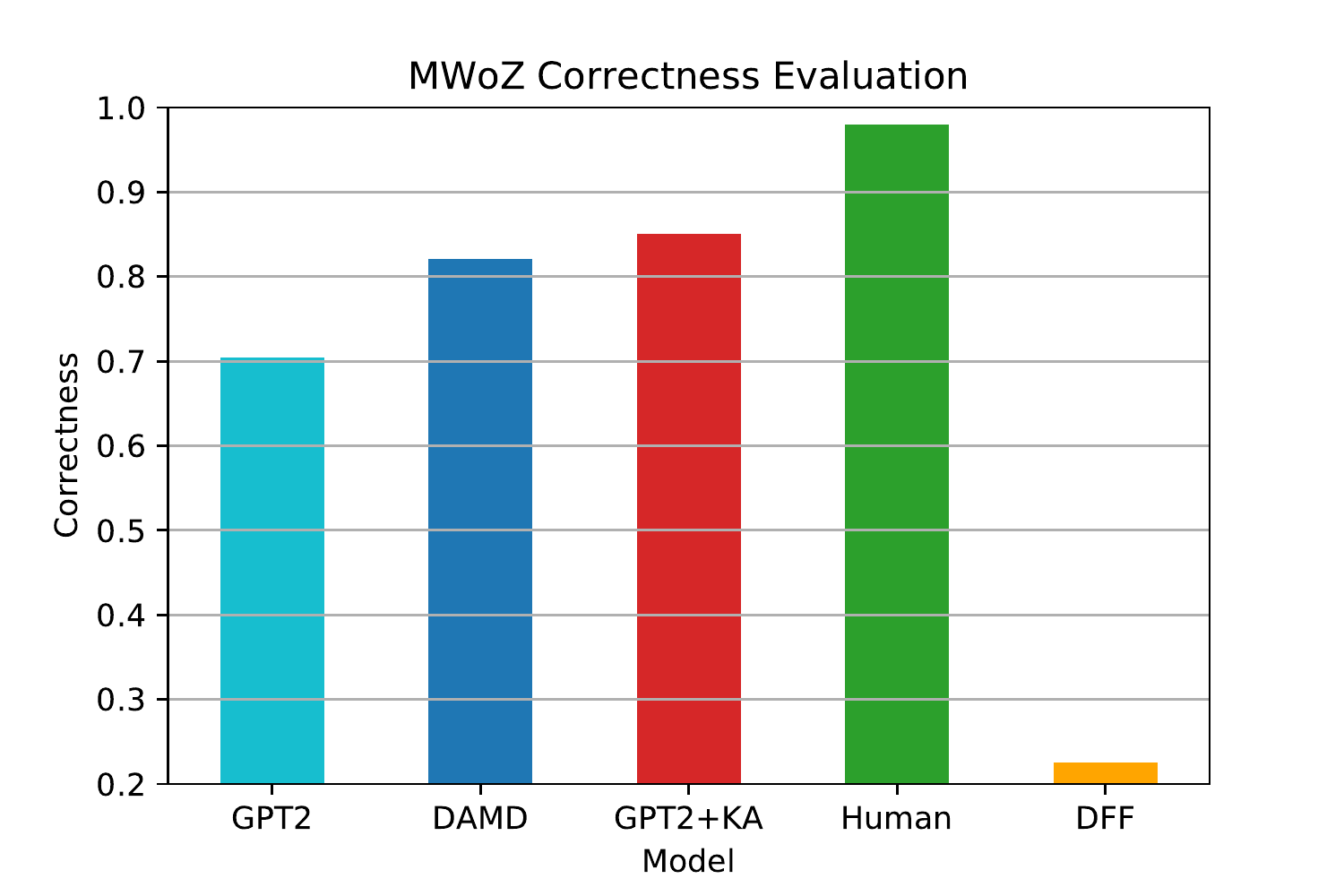}
            \noindent
            \includegraphics[width=0.9\linewidth]{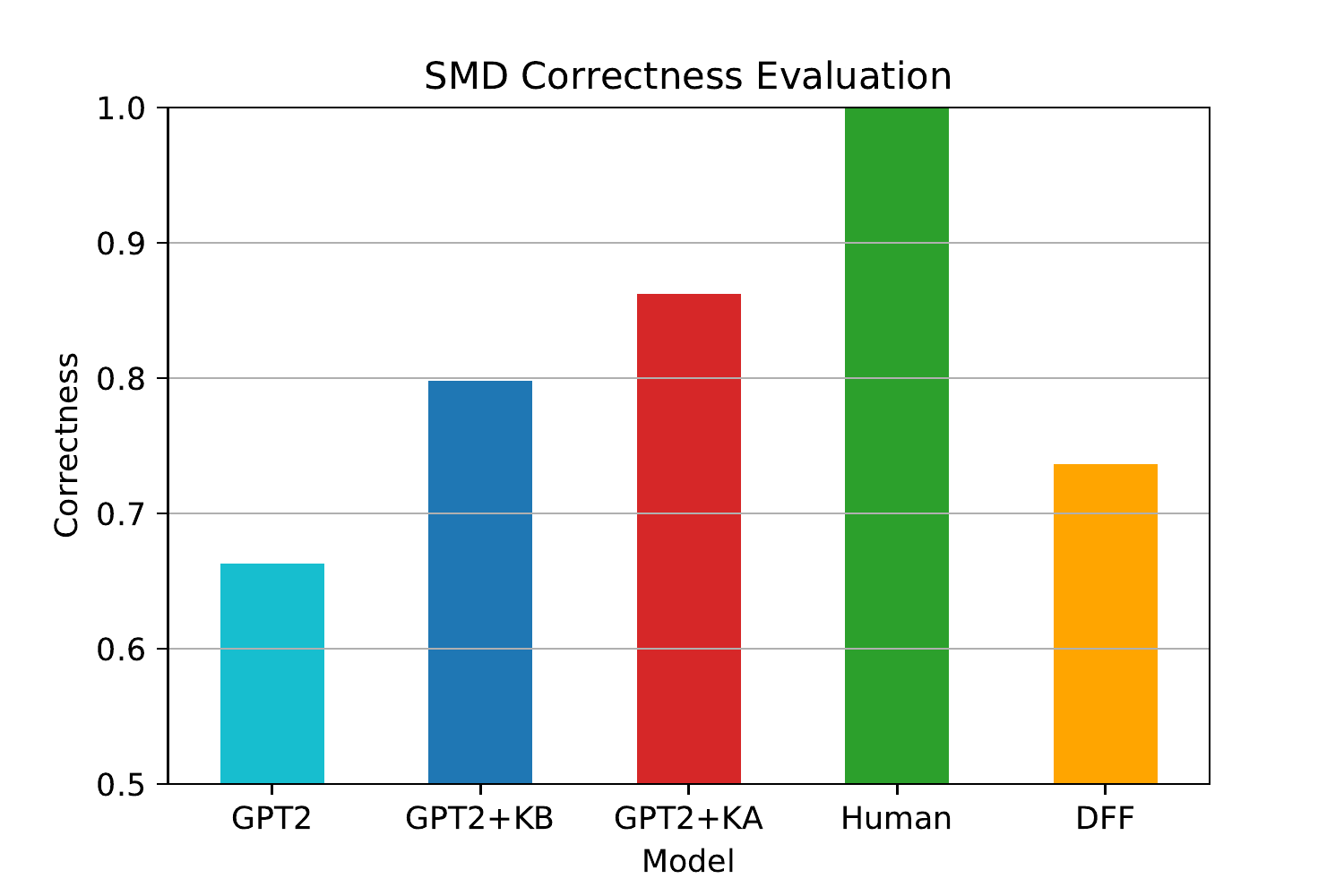}
            \caption{Humanness evaluation in CamRest, MWoZ, and SMD dataset.}
            \label{fig:correctness}
        \endgroup
    \end{minipage}
\end{figure*}

In Humanness collected 3 annotations for each sample, while for correctness we used 1 annotation for each sample made by an expert. We take the mean of the annotation score to get the inter-rater agreement score. Our human evaluation reaches statistical significance with 95\% confidence interval. We report the human evaluation statistics for each dataset in Table \ref{tab:human-evaluation-statistics}. The result of humanness and correctness human evaluation are shown in Figure~\ref{fig:humanness} and Figure~\ref{fig:correctness} respectively.

\begin{table}[t]
\resizebox{0.47\textwidth}{!}{
\begin{tabular}{|c|l|c|c|c|}
\hline
\multicolumn{2}{|c|}{\textbf{Statistics}}  & \textbf{CamRest} & \textbf{SMD} & \textbf{MWoZ} \\ \hline
\multirow{3}{*}{\textbf{Humanness}} & \#annotation & 3  & 3  & 3 \\ \cline{2-5} 
  & \#utterance  & 150  & 450  & 495 \\ \cline{2-5} 
  & avg. deviation & 0.88  & 0.74  & 0.85 \\ \hline
\multicolumn{1}{|l|}{\multirow{2}{*}{\textbf{Correctness}}} & \#annotation & 1  & 1  & 1 \\ \cline{2-5} 
\multicolumn{1}{|l|}{}  & \#utterances & 147  & 255  & 339 \\ \hline
\end{tabular}
}
\caption{Human evaluation statistics.}
\label{tab:human-evaluation-statistics}
\end{table}

\section{System Comparison}
To make a clear distinction of our work to existing task-oriented dialogue systems, we categorize them based on the annotated information and external dependencies used in the pre-processing phase and training-inference phase, such as knowledge base (\textit{KB}), API call for retrieving information(\textit{API}), user goal \textit{Goal}), dialogue span (\textit{Span}), dialogue state tracking (\textit{DST}), speech act (\textit{S-ACT}), and lexicalization response (\textit{LEX-R}). As shown in Table \ref{comparison}, we classify the existing work into four different categories \textit{E2E+Pipelined}, \textit{E2E+API+KB}, \textit{E2E+GOLD KB}, and \textit{E2E+KB}. 

Our work is very distinct to all existing works because our approach does not incorporate any annotated information and external dependencies during training and inference time. Our approach utilizes some annotated information only on the pre-processing phase and it trains the model end-to-end with the knowledge-embedded dataset. Our approach is not only removing the dependencies to external dependencies but also eliminate most of the complexity of the whole training-inference process. 

\section{Experimental Settings}
\label{sec:appendix_experimental_settings}

We report our hyper-parameters to train our model in Table~\ref{tab:smd-hyper} for SMD, CAMREST, and OpenDialKG and Table~\ref{tab:mwoz-hyper} for MultiWOZ 2.1.

\begin{table}[t]
\centering
\resizebox{0.48\textwidth}{!}{
\begin{tabular}{l|ccccc}
\hline
 & \textbf{GPT2} & \multicolumn{1}{l}{\textbf{+KE25}} & \multicolumn{1}{l}{\textbf{+KE50}} &
 \multicolumn{1}{l}{\textbf{+KE75}} &\multicolumn{1}{l}{\textbf{+KE100}} \\ \hline
batch size & 8 & 8 & 8 & 8 & 8 \\
grad accu & 4 & 4 & 4 & 4 & 4 \\
lr & 6.25e-5 & 6.25e-5 & 6.25e-5 & 6.25e-5 & 6.25e-5 \\
epoch & 30 & 30 & 30 & 30 & 30 \\
fp16 & - & - & - & - & - \\
max length & 150 & 150 & 150 & 150 & 150 \\
max history & 50 & 50 & 50 & 50 & 50 \\
num layer & 12 & 12 & 12 & 12 & 12 \\
num head & 12 & 12 & 12 & 12 & 12 \\
num emb & 768 & 768 & 768 & 768 & 768 \\
vocab size & 50k & 50k & 50k & 50k & 50k \\
params & 117M & 117M & 117M & 117M & 117M \\ 
topk & 1 & 1 & 1 & 1 & 1 \\ \hline
\end{tabular}
}
\caption{Hyper-parameters on SMD, CAMREST, and OpenDialKG. The experiments were run on several Nvidia 1080Ti.}
\label{tab:smd-hyper}
\end{table}

\begin{table}[t]
\centering
\resizebox{0.48\textwidth}{!}{
\begin{tabular}{l|cccc}
\hline
 & \textbf{GPT2} & \multicolumn{1}{l}{\textbf{\textbf{+KE25}}} & \multicolumn{1}{l}{\textbf{\textbf{+KE50}}} & \multicolumn{1}{l}{\textbf{\textbf{+KE100}}} \\ \hline
batch size & 6 & 6 & 6 & 6 \\
grad accu & 3 & 3 & 3 & 3 \\
lr & 6.25e-5 & 6.25e-5 & 6.25e-5 & 6.25e-5 \\
epoch & 10 & 10 & 10 & 5 \\
fp16 & O2 & O2 & O2 & O2 \\
max length & 150 & 150 & 150 & 150 \\
max history & 50 & 50 & 50 & 50 \\
num layer & 12 & 12 & 12 & 12 \\
num head & 12 & 12 & 12 & 12 \\
num emb & 768 & 768 & 768 & 768 \\
vocab size & 50k & 50k & 50k & 50k \\
params & 117M & 117M & 117M & 117M \\ 
topk & 1 & 1 & 1 & 1 \\ \hline
\end{tabular}
}
\caption{Hyper-parameters on MultiWOZ. The experiments were run on a single Nvidia V100.}
\label{tab:mwoz-hyper}
\end{table}


\section{Datasets Information}
Table \ref{fig:dataset-stats} shows the data splits (train/valid/test) and the link to download each dataset.
\begin{table}[t]
\centering
\resizebox{0.48\textwidth}{!}{
\begin{tabular}{l|ccc|l}
\hline
\multirow{2}{*}{\textbf{Dataset}} & \multicolumn{3}{c|}{\textbf{Split}} & \multirow{2}{*}{\textbf{Source}} \\ \cline{2-4}
 & \multicolumn{1}{c}{\textbf{Train}} & \multicolumn{1}{c}{\textbf{Valid}} & \multicolumn{1}{c|}{\textbf{Test}} & \\ \hline
bAbI & 1,000 & 1,000 & 1,000 & \hyperlink{https://research.fb.com/downloads/babi/}{Website}\\ \hline
CAMREST & 406 & 135 & 135 & \hyperlink{https://github.com/dair-iitd/BossNet}{Github repository} \\ \hline
SMD (KVR) & 2,425 & 302 & 304 & \hyperlink{http://nlp.stanford.edu/projects/kvret/kvret_dataset_public.zip}{Website} \\ \hline
MultiWOZ & 2,447 & 204 & 226 & \multirow{6}{*}{\hyperlink{https://github.com/budzianowski/multiwoz/raw/master/data/MultiWOZ_2.1.zip}{Github repository}} \\
\hspace{2mm}attraction single & 127 & 11 & 12 & \\
\hspace{2mm}hotel single & 513 & 56 & 67 & \\
\hspace{2mm}restaurant single & 1,199 & 50 & 62 & \\
\hspace{2mm}taxi single & 326 & 57 & 52 & \\
\hspace{2mm}train single & 282 & 30 & 33 & \\ \hline
OpenDialKG & 11,041 & 1,380 & 1,380 & \hyperlink{https://github.com/facebookresearch/opendialkg.git}{Facebook Github repository} \\ \hline
\end{tabular}
}
\caption{Dataset Statistics and Source.}
\label{fig:dataset-stats}
\end{table}

\section{Detailed Experiment Results}

We report more detailed results for bAbI-5, SMD, CamRest and MwoZ. Figure~\ref{tab:babi-results-app} shows all detailed results in bAbI dataset. Figure~\ref{tab:SMD-app} shows all detailed results in SMD dataset. Figure~\ref{tab:detail_CamRest} shows all detailed results on CamRest676 dataset.  Figure~\ref{tab:detail_MWoZ} shows all detailed results on MWoZ 2.1 dataset.

\begin{table}[t]
\centering
\resizebox{0.47\textwidth}{!}{
\begin{tabular}{r|c|c}
\hline
\multirow{1}{*}{\textbf{Model}} & \multicolumn{1}{c|}{\textbf{Test}} & \multicolumn{1}{c}{\textbf{Test OOV}} \\ \hline
QRN$^1$ & 99.60 (-) & 67.80 (-) \\
Mem2Seq$^2$ & 97.90 (69.60) & 84.50 (2.30)  \\
BoSsNet$^3$ & 97.30 (65.60) & 91.70 (18.50)\\
GLMP$^4$ & 99.20 (88.50) & 92.00 (21.70)  \\ \hline
GPT2 & 90.74 (31.00) & 70.14 (0.00) \\ 
GPT2+KE 1& 93.31 (46.10) & 74.75 (2.00)\\
GPT2+KE 10& 99.84 (98.10) & 96.84 (77.20) \\
GPT2+KE 50& 99.78 (97.10) & \textbf{99.60} (\textbf{95.70}) \\
GPT2+KE 100& \textbf{99.99} (\textbf{99.90}) & 99.01 (94.90) \\ \hline
\end{tabular}
}
\caption{Results on the bAbI dataset.$^1$~\cite{seo2017query}, $^2$~\cite{madotto2018mem2seq}, $^3$~\cite{raghu2019disentangling}, $^3$~\cite{wu2019global}.}
\label{tab:babi-results-app}
\end{table}

\begin{table}[t]
\centering
\resizebox{0.47\textwidth}{!}{
\begin{tabular}{r|ccc|cc}
\hline
\textbf{Model} & \textbf{Success} & \textbf{BLEU} & \textbf{F1} & \textbf{Human} & \textbf{Correct} \\ \hline
Human & 86.08 & - & - & 3.60 & 96.97 \\ \hline
KB-Trs$^1$ & - &  14.80 & 45.30 & - & - \\
MLMN$^2$ & - & 13.61 & 54.85 & - & - \\ 
BoSsNet$^3$ & - & 15.20 & 43.10 & - & - \\
KBRet$^4$ & 62.03 & \textbf{18.64} & 55.76 & 3.13 & 77.33 \\ 
\hline
GPT2 & 30.38 & 13.58 & 34.69 & 3.42 & 66.67 \\
GPT2+KB & 62.03 & 13.59 & 50.45 & 2.42 & 70.37 \\
GPT2+KE10 & 62.03 & 16.55 & 52.15 & - & - \\
GPT2+KE50 & 70.89 & 17.85 & \textbf{55.81}  & - & - \\
GPT2+KE100 & 72.15 & 17.78 & 54.04 & - & - \\ 
GPT2+KE161 & \textbf{74.68} & \textbf{18.00} & 54.85 & \textbf{3.48} & \textbf{83.50} \\ \hline
\end{tabular}   
}
\caption{Detailed results on CAMREST dataset. $^1$\cite{haihong2019kb}. $^2$\cite{reddy2019multi}.
$^3$\cite{raghu2019disentangling}.
$^4$\cite{qin2019entity}. We re-evaluate $^4$ using our script that includes postcode as entity and removes the API-call from the F1-count.}
\label{tab:detail_CamRest}
\end{table}

\begin{table*}[t]
\centering
\resizebox{0.67\textwidth}{!}{
\begin{tabular}{r|c|cccc|cc}
\hline
\textbf{Model} & \textbf{BLEU} & \textbf{Ent.} & \textbf{Nav.} & \textbf{Wea.} & \textbf{Sch.} & \textbf{Hum.} & \textbf{Cor.}  \\ \hline
KVRet$^1$ & 13.20 & 48.00 & 44.50 & 53.30 & 62.90 & - & - \\ 
MLMN$^2$ & 17.10 & 55.10 & 41.30 & 47.00 & 68.30 & - & - \\\hline
BoSsNet$^3$ & 8.3 & 35.9 & - & - & - & - & - \\
Mem2Seq$^4$ & 12.20 & 33.40 & 20.00 & 49.30 & 32.80 & - & - \\
KBRet$^5$ & 13.90 & 53.70 & 54.50 & 52.20 & 55.60 & - & - \\
KB-Trs$^6$ & 13.90 & 37.10 &  23.30 & 48.20 & 51.20 & - & - \\
GLMP$^7$ & 13.90 & 60.70 & 54.60 & 56.50 & 72.50 & - & - \\
DFF$^8$ & 14.40 & \textbf{62.70} & \textbf{57.90} & 57.60 & \textbf{73.10} & 3.28 & 68.90 \\ \hline
GPT2 & 15.60 & 39.11 & 23.41 & 53.74 & 52.26 & \textbf{3.49} & 67.05 \\
GPT2+KB & 17.03 & 58.60 & 48.37 & \textbf{62.87} & 72.22 & 3.47 & 81.03 \\
GPT2+KE 10 & 14.18 & 52.88 & 50.26 & 51.64 & 58.62 & - & - \\ 
GPT2+KE 25 & 14.22 & 55.00 & 50.46 & 52.91 & 64.87 & - & - \\ 
GPT2+KE 50 & 14.90 & 56.43 & 50.04 & 54.25 & 69.60 & - & - \\ 
GPT2+KE 75 & 16.31 & 58.79 & 52.56 & 56.39 & 71.89 & - & - \\
GPT2+KE 100 & \textbf{17.35} & \textbf{59.78} & \textbf{53.53} & 57.73 & \textbf{72.58} & 3.44 & \textbf{85.56} \\ \hline
Human$^1$ & 13.50 & 60.70 & 55.20 & 61.60 &  64.30 & 3.54 &  97.92 \\ \hline
\end{tabular} 
}
\caption{Results on the SMD (KVR) dataset. $^{^1}$\citet{eric2017key} $^2$\cite{reddy2019multi}
$^3$\cite{raghu2019disentangling} $^4$\cite{madotto2018mem2seq} $^5$\cite{qin2019entity} $^6$\cite{haihong2019kb} $^7$\cite{wu2019global} $^8$\cite{qin2020dynamic}} 
\label{tab:SMD-app}
\end{table*}

\begin{table*}[t]
\resizebox{0.98\textwidth}{!}{
\begin{tabular}{r|cc|cc|ccccc|cc}
\hline
\textbf{Model}   &
\multicolumn{1}{c}{\textbf{Inform}} & \multicolumn{1}{c|}{\textbf{Success}} & \multicolumn{1}{c}{\textbf{BLEU}} & \multicolumn{1}{c|}{\textbf{F1}} & \multicolumn{1}{c}{\textbf{Train}} & \multicolumn{1}{c}{\textbf{Attraction}} & \multicolumn{1}{c}{\textbf{Hotel}} & \multicolumn{1}{c}{\textbf{Rest}} & \multicolumn{1}{c|}{\textbf{Taxi}} & \multicolumn{1}{c}{\textbf{Human}} & \multicolumn{1}{c}{\textbf{Correct}} \\ \hline
Human & - & - & - &  - & - & - & - & - & - & 3.66 & 96.85 \\ \hline
Mem2Seq$^2$ & - & - & 6.60 &  21.62 & - & 22.00 & 21.00 & 22.40 & - & - & - \\
DSR$^3$ & - & - & 9.10 & - & 30.00 & 28.00 & 27.00 & 33.40 & - & - & - \\
GLMP$^4$ & - & - & 6.90 & - & 32.40 & 24.40 & 28.10 & 38.40 & - & - & - \\
DFF$^5$ & - & - & 9.40 & - & 35.10 & 28.10 & 30.60 & \textbf{40.90} & - & 2.65 & 25.53 \\ \hline
GPT2 & 64.60 & 51.77 & 14.33 & 30.38 & 23.30  &  15.11  &  23.56  &  25.62  & 89.76 & 3.51 & 55.91 \\
GPT2+KE-25 &  70.80  &  57.52  & 14.24 & 36.96 & 22.27  &  43.30  &  29.74  &  35.71  & 87.62 & - & - \\
GPT2+KE-50 & 72.12  &  58.41  & 13.44 & 37.20 &  21.95  &  \textbf{44.72}  &  30.03  &  36.10  & 87.38 & - & -  \\
GPT2+KE-100 & \textbf{72.57}  &  \textbf{64.16}  & \textbf{15.05}  &  \textbf{39.58} &  \textbf{23.79}  &  43.32  &  \textbf{33.44}  &  37.10  & \textbf{92.38} & \textbf{3.56} & \textbf{73.38} \\ \hline
DAMD$^1$ & 85.40 & 70.40 & 13.50 & -& -&- &- &-&- &- &- \\
DAMD$^{\star}$ & 72.12  &  61.06 & 11.48 & 22.58  & 16.96 & 31.05 & 15.50 & 22.23 & 55.95 & 3.31 & 67.97 \\ \hline
\end{tabular}
}
\caption{Detailed results on MultiWOZ dataset. $^1$\cite{zhang2019task}. $^2$\cite{madotto2018mem2seq}. $^3$\cite{wen2018sequence}. $^4$\cite{wu2019global}. $^5$\cite{qin2020dynamic}. $^{\star}$We evaluate DAMD with our scorer.}
\label{tab:detail_MWoZ}
\end{table*}

\section{How many \texttt{TEMPLATE}s are enough?}
\begin{figure*}[t]
    \centering
    \includegraphics[scale=0.16]{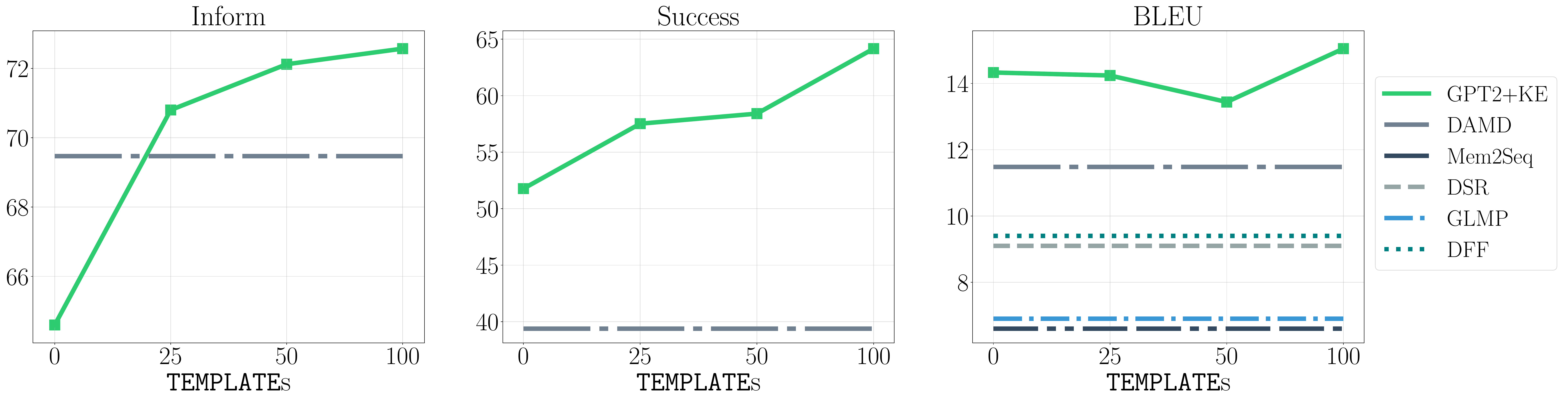}
    \caption{Inform, Success, BLEU score versus number of \texttt{TEMPLATE}s in the MultiWOZ dataset.} \label{fig:MWoZ_by_template}
    \vspace{20pt}
    \includegraphics[scale=0.14]{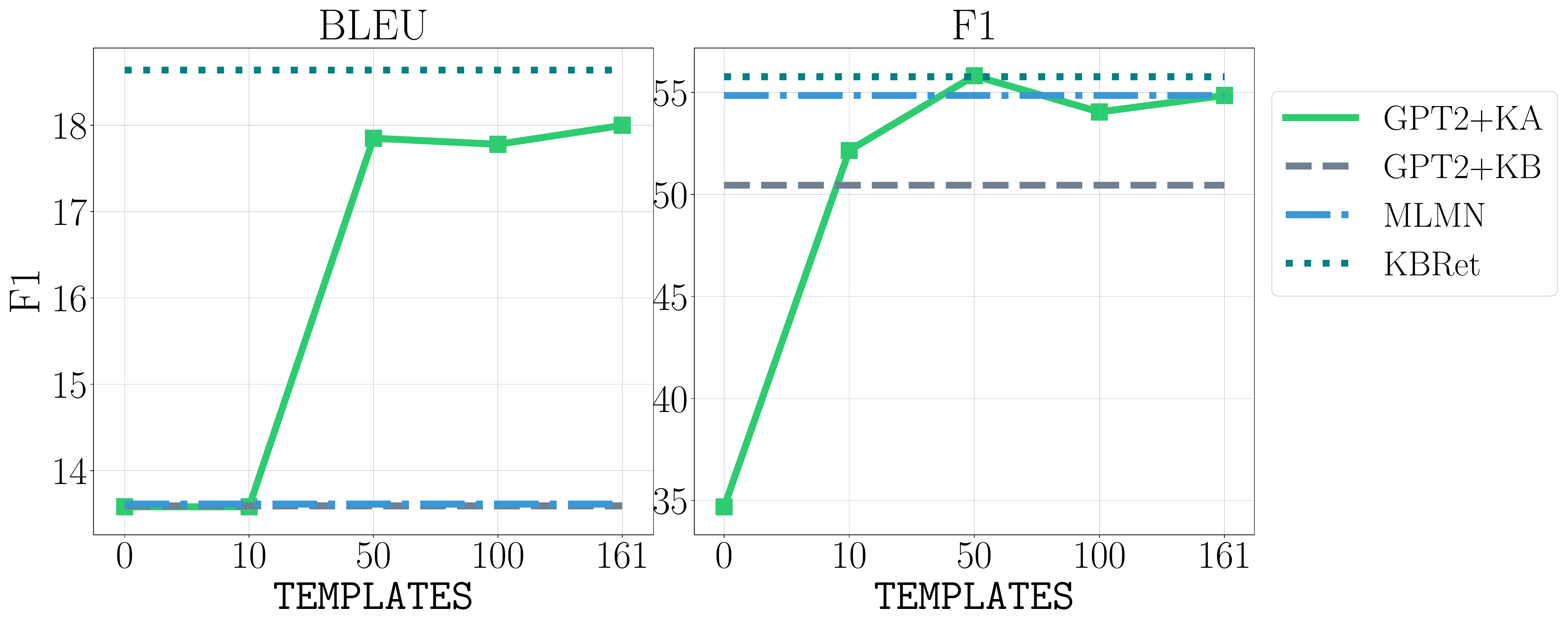}
    \caption{BLEU and F1-Score versus number of \texttt{TEMPLATE}s by domain in the CamRest dataset.}     \label{fig:CamRest_by_template}
    \vspace{20pt}
    \includegraphics[scale=0.14]{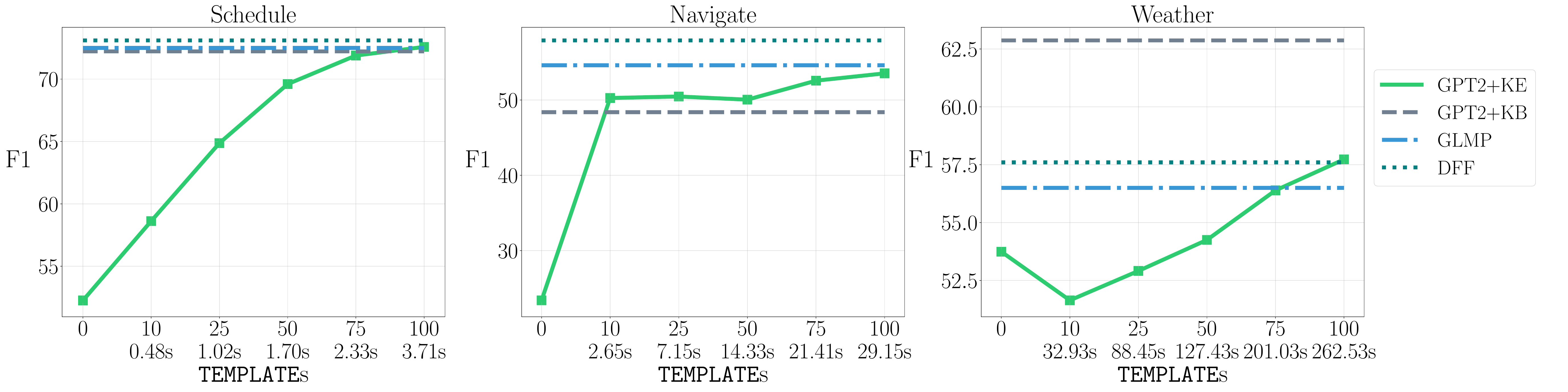}
    \caption{F1-Score versus number of \texttt{TEMPLATE}s by domain in the SMD~\cite{ericKVR2017} dataset. In the x-axes we report the average fine-tuning time per epoch. } 
    \label{fig:SMD_by_template}
\end{figure*}

We further analyze our result to see how many \texttt{TEMPLATE}s are enough to achieve good performance in the corresponding dataset. In CamRest dataset, as shown in Figure~\ref{fig:CamRest_by_template}, we can see that there is a steep increase from without KE-dialogue to 10 \texttt{TEMPLATE}s in term of F1 and a steep improvement from 10 \texttt{TEMPLATE}s to 50 \texttt{TEMPLATE}s in term of BLEU. This fact suggests that 50 \texttt{TEMPLATE}s on CamRest dataset is enough to represent the whole dataset. In MWoZ dataset, as shown in Figure \ref{fig:MWoZ_by_template}, with 100 templates the inform and success scores are still increasing while the BLEU score remains stable over \texttt{TEMPLATE}s. This suggests that we need more than 100 \texttt{TEMPLATE}s to get the optimum benefit from our approach. 

In SMD dataset, as shown in  \ref{fig:SMD_by_template}, in Schedule domain the F1-scores keep increasing steadily until 50 \texttt{TEMPLATE}s and slowing down in 75 and 100 \texttt{TEMPLATE}s. In Navigation domain there is a steep increase of F1-score from the one without KE-dialogue to the one with 10 \texttt{TEMPLATE}s. In weather domain, the F1-score increases steadily from 10 to 100 \texttt{TEMPLATE}s. This results suggest on Schedule domain, around 100 \texttt{TEMPLATE} is needed to get the optimal score, while on navigation domain, only a around 10 to 25 \texttt{TEMPLATE}s is required, and Weather domain more than 100 \texttt{TEMPLATE}s is required in order to achieve the optimal score.

\section{Example of Template Generation}
\label{template_generation}

Examples illustrating the \texttt{KE-DELEX} and \texttt{KE-RELEX} process for MultiWOZ, CamRest, and SMD datasets are, respectively, shown in Table~\ref{tab:de-relex-multiwoz}, Table~\ref{tab:de-relex-cam}, and Table~\ref{tab:de-relex-smd}. For CamRest dataset, we remove all of the API calls in the dialogue. For MultiWOZ we keep booking API call and keep the booking reference number as is.

\begin{table*}[t]
\centering
\begin{minipage}{.99\linewidth}
\resizebox{\textwidth}{!}{
\begin{tabular}{l|p{10cm}|p{10cm}}
\hline
\texttt{TURN} & \multicolumn{1}{c}{\texttt{ORIGINAL TEXT}} & \multicolumn{1}{|c}{\texttt{TEMPLATE}} \\ \hline
USR & i am looking for a particular restaurant . it is called pizza hut city centre . & i am looking for a particular restaurant . it is called \textbf{[name\_0]} . \\
SYS & it s a cheap italian place in the centre at regent street city centre cb21ab . the phone number is 01223323737 . would you like more info ? & it s a \textbf{[price\_0]} \textbf{[food\_0]} place in the \textbf{{[area\_0]}} at \textbf{{[address\_0]}} \textbf{{[post\_0]}} . the phone number is \textbf{{[phone\_0]}} . would you like more info ? \\
USR & that sounds like just the place ! can you book me a table for 2 at 1945 on thursday ? & that sounds like just the place ! can you book me a table for 2 at 1945 on thursday ? \\
SYS-API & restaurant name = 'pizza hut city centre' time = '1945' day = 'thursday' people = '2' & restaurant name = '\textbf{{[name\_0]}}' time = '1945' day = 'thursday' people = '2' \\
API & \multicolumn{1}{l|}{f3k2pqzz} & \multicolumn{1}{l|}{f3k2pqzz}  \\
SYS & you are all set . i have got a table for 2 at 1945 this thursday . your reference number is f3k2pqzz . & you are all set . i have got a table for 2 at 1945 this thursday . your reference number is f3k2pqzz . \\
USR & ok sounds good . thank you ! & ok sounds good . thank you ! \\
SYS & you are welcome . is there anything else i can assist you with today ? & you are welcome . is there anything else i can assist you with today ? \\
USR & that is all . thank you for your help ! & that is all . thank you for your help ! \\
SYS & you are welcome . have a great day . & you are welcome . have a great day . \\ \hline
\multicolumn{3}{l}{} \\ \hline
\multicolumn{3}{l}{$\texttt{USER GOAL QUERY}$} \\
\multicolumn{3}{l}{\textbf{SELECT} area, food, price, name, address, phone, postcode \textbf{FROM} Restaurant} \\ \hline 
\end{tabular}
}
\end{minipage}
\begin{minipage}{.99\linewidth}
\resizebox{\textwidth}{!}{
\begin{tabular}{ccccccc}
\\ \hline
\textbf{Area} & \textbf{Food} & \textbf{Price} & \textbf{Name} & \textbf{Address} & \textbf{Phone} & \textbf{Postcode} \\ \hline
centre & italian & cheap & pizza hut city centre & Regent Street City Centre & 01223323737 & cb21ab \\
centre & indian & cheap & kohinoor & 74 Mill Road City Centre & 01223323639 & cb12as \\
centre & chinese & \multicolumn{1}{c}{expensive} & hk fusion & 21 Burleigh Street City Centre & 01223355909 & cb11dg \\
south & \multicolumn{1}{c}{modern european} & \multicolumn{1}{c}{moderate} & restaurant alimentum & 152 - 154 Hills Road & 01223413000 & cb28pb \\ \hline \\
\end{tabular}
}
\end{minipage}
\begin{minipage}{.99\linewidth}
\resizebox{\textwidth}{!}{
\begin{tabular}{l|p{10cm}|p{10cm}} \hline
\texttt{TURN} & \multicolumn{1}{c}{\texttt{RELEX 1}} & \multicolumn{1}{|c}{\texttt{RELEX 2}} \\ \hline
USR & i am looking for a particular restaurant . it is called \textbf{kohinoor} . & i am looking for a particular restaurant . it is called \textbf{restaurant alimentum} . \\
SYS & it s a \textbf{cheap indian} place in the \textbf{centre} at \textbf{74 mill road city centre} \textbf{cb12as} . the phone number is \textbf{01223323639} . would you like more info ? & it s a \textbf{moderate} \textbf{modern european} place in the \textbf{south} at \textbf{152 - 154 hills road} \textbf{cb28pb} . the phone number is \textbf{01223413000} . would you like more info ? \\
USR & that sounds like just the place ! can you book me a table for 2 at 1945 on thursday ? & that sounds like just the place ! can you book me a table for 2 at 1945 on thursday ? \\
SYS-API & restaurant name = '\textbf{kohinoor}' time = '1945' day = 'thursday' people = '2' & restaurant name = '\textbf{restaurant alimentum}' time = '1945' day = 'thursday' people = '2' \\
API & \multicolumn{1}{l|}{f3k2pqzz} & \multicolumn{1}{l|}{f3k2pqzz} \\
SYS & you are all set . i have got a table for 2 at 1945 this thursday . your reference number is f3k2pqzz . & you are all set . i have got a table for 2 at 1945 this thursday . your reference number is f3k2pqzz . \\
USR & ok sounds good . thank you ! & ok sounds good . thank you ! \\
SYS & you are welcome . is there anything else i can assist you with today ? & you are welcome . is there anything else i can assist you with today ? \\
USR & that is all . thank you for your help ! & that is all . thank you for your help ! \\
SYS & you are welcome . have a great day . & you are welcome . have a great day . \\ \hline
\end{tabular}
}
\end{minipage}
\caption{MultiWOZ 2.1 \texttt{KE-DELEX} and \texttt{KE-RELEX} Examples}
\label{tab:de-relex-multiwoz}
\end{table*}

\begin{table*}[t]
\centering
\begin{minipage}{.99\linewidth}
\resizebox{\textwidth}{!}{
\begin{tabular}{l|p{9cm}|p{9cm}}
\hline
\multicolumn{1}{c|}{\texttt{TURN}} &
  \multicolumn{1}{c|}{\texttt{ORIGINAL TEXT}} &
  \multicolumn{1}{c}{\texttt{TEMPLATE}} \\ \hline
USR &
  i want a moderately priced restaurant in the east part of town . &
  i want a \textbf{[price\_1]}ly priced \textbf{[type\_1]} in the \textbf{[area\_1]} part of town . \\
SYS &
  would you prefer indian food or italian food ? &
  would you prefer \textbf{[food\_2]} food or \textbf{[food\_3]} food ? \\
USR &
  i do n't care . give me their phone number . &
  i do n't care . give me their phone number . \\
SYS &
  curry prince is a moderately priced restaurant in the east part of town that serves indian food . would you like the phone number ? &
  \textbf{[name\_2]} is a \textbf{[price\_2]}ly priced \textbf{[type\_2]} in the \textbf{[area\_2]} part of town that serves \textbf{[food\_2]} food . would you like the phone number ? \\
USR &
  yes give me the phone number . &
  yes give me the phone number . \\
SYS &
  the phone number for Curry Prince is 01223566388 . &
  the phone number for \textbf{[name\_2]} is \textbf{[phone\_2]} . \\
USR &
  awesome , thank you ! goodbye ! &
  awesome , thank you ! goodbye ! \\
SYS &
  you 're welcome , goodbye . &
  you 're welcome , goodbye . \\ \hline
\multicolumn{3}{l}{} \\ \hline
\multicolumn{3}{l}{$\texttt{USER GOAL QUERY}$} \\
\multicolumn{3}{l}{\textbf{SELECT} area, food, price, name, phone \textbf{FROM} Restaurant} \\ \hline 
\end{tabular}
}
\end{minipage}
\begin{minipage}{.99\linewidth}
\resizebox{\textwidth}{!}{
\begin{tabular}{cccccccc}
\\ \hline
\textbf{Area} & \textbf{Food} & \textbf{Price} & \textbf{Name} & \textbf{Address} & \textbf{Phone} & \textbf{Postcode} & \textbf{Location} \\ \hline
east & indian & moderate & curry prince & 451 newmarket road & 01223566388 & cb58jj & 52.213072,0.149771\\
east & italian & moderate & pizza hut & cambridge retail park & 01223323737 & cb58wr & 52.209742,0.146975\\
east & indian & moderate & rajmahal & 7 barnwell road & 01223244955 & cb58rg & 52.211144,0.16324 \\ \hline \\
\end{tabular}
}
\end{minipage}
\begin{minipage}{.99\linewidth}
\resizebox{\textwidth}{!}{
\begin{tabular}{l|p{9cm}|p{9cm}}
\hline
\multicolumn{1}{c|}{\texttt{TURN}} &
  \multicolumn{1}{c|}{\texttt{RELEX 1}} &
  \multicolumn{1}{c}{\texttt{RELEX 2}} \\ \hline
USR &
  i want a \textbf{moderate}ly priced \textbf{restaurant} in the \textbf{east} part of town . &
  i want a \textbf{moderate}ly priced \textbf{restaurant} in the \textbf{east} part of town . \\
SYS &
  would you prefer \textbf{indian} food or \textbf{italian} food ? &
  would you prefer \textbf{italian} food or \textbf{indian} food ? \\
USR &
  i do n't care . give me their phone number . &
  i do n't care . give me their phone number . \\
SYS &
  \textbf{rajmahal} is a \textbf{moderate}ly priced \textbf{restaurant} in the \textbf{east} part of town that serves \textbf{indian} food . would you like the phone number ? &
  \textbf{pizza hut} is a \textbf{moderate}ly priced \textbf{restaurant} in the \textbf{east} part of town that serves \textbf{italian} food . would you like the phone number ? \\
USR &
  yes give me the phone number . &
  yes give me the phone number . \\
SYS &
  the phone number for \textbf{rajmahal} is \textbf{01223244955} . &
  the phone number for \textbf{pizza hut} is \textbf{01223323737} . \\
USR &
  awesome , thank you ! goodbye ! &
  awesome , thank you ! goodbye ! \\
SYS &
  you 're welcome , goodbye . &
  you 're welcome , goodbye . \\ \hline 
\end{tabular}
}
\end{minipage}
\caption{CamRest Delex and Relex Examples}
\label{tab:de-relex-cam}
\end{table*}

\begin{table*}[!t]
\begin{minipage}{\linewidth}
\centering
\resizebox{0.85\textwidth}{!}{
\begin{tabular}{l|l|l}
\hline
\texttt{TURN} & \multicolumn{1}{c|}{\texttt{ORIGINAL TEXT}} &  \multicolumn{1}{c}{\texttt{TEMPLATE}} \\ \hline
USR & where is the closest rest stop ?  & where is the closest \textbf{[poi\_type\_1]} ?      \\
SYS & the westin is 2 miles away  & \textbf{[poi\_11]} is \textbf{[distance\_11]} away            \\
USR & what is the address ?  & what is the address ?                     \\
SYS & the westin is located at 329 el camino real & \textbf{[poi\_11]} is located at \textbf{[poi\_address\_11]} .  \\
USR & thanks set the navigation to there please  & thanks set the navigation to there please \\
SYS & you are welcome , i set the navigation     & you are welcome , i set the navigation    \\
\hline
\multicolumn{3}{l}{} \\ \hline 
\multicolumn{3}{l}{\texttt{User Goal Query}} \\ \multicolumn{3}{l}{\begin{tabular}[c]{@{}l@{}}\textbf{SELECT} type, poi, dist, addr \textbf{FROM} navigation \textbf{GROUP BY} type \textbf{HAVING} dist = \textbf{MIN}(dist)\end{tabular}} \\
\hline
\multicolumn{3}{l}{} \\
\end{tabular}
}
\end{minipage}
\begin{minipage}{\linewidth}
\centering
\resizebox{0.85\textwidth}{!}{
\begin{tabular}{ccccc}
\hline
\multicolumn{1}{c}{\textbf{POI}} &
  \multicolumn{1}{c}{\textbf{POI Type}} &
  \multicolumn{1}{c}{\textbf{Address}} &
  \multicolumn{1}{c}{\textbf{Distance}} &
  \multicolumn{1}{c}{\textbf{Traffic}} \\ \hline
whole foods             & grocery store      & 819 alma st         & 2 miles & heavy traffic \\
hotel keen              & rest stop          & 578 arbol dr        & 3 miles & no traffic    \\
home & home    & 338 alester ave     & 3 miles & no traffic    \\
round table             & pizza restaurant   & 113 anton ct        & 4 miles & heavy traffic \\
mandarin roots          & chinese restaurant & 271 springer street & 3 miles & heavy traffic \\ \hline
\multicolumn{5}{l}{} \\ 
\end{tabular}
}
\end{minipage}
\begin{minipage}{\linewidth}
\centering
\resizebox{0.85\textwidth}{!}{
\begin{tabular}{l|l|l}
\hline
\texttt{TURN} & \multicolumn{1}{c|}{\texttt{RELEX 1}} & \multicolumn{1}{c}{\texttt{RELEX 2}}  \\ \hline
USR & where is the closest \textbf{grocery store} ? & where is the closest \textbf{home} ?      \\
SYS & \textbf{whole foods} is \textbf{2 miles} away& \textbf{home} is \textbf{3 miles} away            \\
USR & what is the address ?     & what is the address ?                     \\
SYS & \textbf{whole foods} is located at \textbf{819 alma st}& \textbf{home} is located at \textbf{338 alester ave}  \\
USR & thanks set the navigation to there please  & thanks set the navigation to there please \\
SYS & you are welcome , i set the navigation     & you are welcome , i set the navigation    \\ 
\hline
\end{tabular}
}
\end{minipage}
\caption{SMD Delex and Relex Example}
\label{tab:de-relex-smd}
\end{table*}

\end{document}